\documentclass{article}

\usepackage{microtype}
\usepackage{graphicx}
\usepackage{subfigure}
\usepackage{booktabs} %

\usepackage{natbib}

\usepackage[nosumlimits,nonamelimits]{amsmath}
\usepackage{amssymb}
\usepackage[pdftex]{hyperref}
\usepackage{url}            %
\usepackage[capitalise]{cleveref}
\usepackage{url}
\usepackage{babel} 
\usepackage{siunitx}
\usepackage{graphicx}
\graphicspath{{figs/}}
\usepackage{printlen}
\usepackage{layouts}
\usepackage{dsfont}
\usepackage{physics}
\usepackage{mathtools}
\usepackage{enumitem}
\usepackage{algpseudocode}
\usepackage{xcolor}         %
\usepackage{tikz}
\usepackage{tcolorbox}

\usepackage{csquotes}
\usepackage[accepted]{icml2021}

\newif\ifarxiv
\arxivtrue 

\newif\ifvspace
\vspacefalse

\usepackage[shortcuts]{extdash}

\usepackage{xfrac}

\icmltitlerunning{Active Testing: Sample--Efficient Model Evaluation}
\usepackage{placeins}
\begin{document}
\newcommand{\seb}[1]{\textcolor{red}{Seb: #1}}
\newcommand{\Dt}{\mathcal{D}_\mathrm{test}}
\newcommand{\Dto}{\mathcal{D}_\mathrm{test}^\mathrm{observed}}
\newcommand{\Dtu}{\mathcal{D}_\mathrm{test}^\mathrm{unobserved}}
\newcommand{\Dtr}{\mathcal{D}_{\textup{train}}}

\newcommand{\Rf}{\hat{R}_{\textup{LURE}}}
\newcommand{\Riid}{\hat{R}_{\textup{iid}}}

\newcommand{\Ls}{\mathcal{L}}
\newcommand{\xin}{\mathbf{x}}
\newcommand{\yin}{\mathbf{y}}
\newcommand{\E}[2]{\mathbb{E}_{#1}\left[#2\right]}
\newcommand{\Var}[2]{\mathbb{V}_{#1}\left[#2\right]}
\newcommand*\circled[1]{\tikz[baseline=(char.base)]{
		\node[shape=circle,draw,inner sep=2pt] (char) {#1};}}
\newcommand{\approptoinn}[2]{\mathrel{\vcenter{
  \offinterlineskip\halign{\hfil$##$\cr
    #1\propto\cr\noalign{\kern2pt}#1\sim\cr\noalign{\kern-2pt}}}}}

\newcommand{\appropto}{\mathpalette\approptoinn\relax}
\newcommand{\surr}[1]{p_{\textup{surr}}(#1\nolinebreak\mid\nolinebreak\xin)}

\definecolor{pal0}{rgb}{0.00, 0.45, 0.70}
\definecolor{pal1}{rgb}{0.87, 0.56, 0.02}
\definecolor{pal2}{rgb}{0.01, 0.62, 0.45}
\definecolor{pal3}{rgb}{0.84, 0.37, 0.00}
\definecolor{pal4}{rgb}{0.80, 0.47, 0.74}
\definecolor{pal5}{rgb}{0.65, 0.34, 0.16}
\definecolor{pal6}{rgb}{0.97, 0.51, 0.75}
\definecolor{pal7}{rgb}{0.58, 0.58, 0.58}
\definecolor{pal8}{rgb}{0.87, 0.87, 0.00}
\definecolor{pal9}{rgb}{0.34, 0.71, 0.91}

\twocolumn[
\icmltitle{Active Testing: Sample--Efficient Model Evaluation}

\icmlsetsymbol{equal}{*}

\begin{icmlauthorlist}
\icmlauthor{Jannik Kossen}{equal,oatml}
\icmlauthor{Sebastian Farquhar}{equal,oatml}
\icmlauthor{Yarin Gal}{oatml}
\icmlauthor{Tom Rainforth}{stats}
\end{icmlauthorlist}

\icmlaffiliation{oatml}{OATML, Department of Computer Science,}
\icmlaffiliation{stats}{Department of Statistics, Oxford}

\icmlcorrespondingauthor{Jannik Kossen}{jannik.kossen@cs.ox.ac.uk}

\icmlkeywords{Machine Learning, ICML, Active Learning, Active Testing, Monte Carlo, Sample-Efficient, Uncertainty, Acquisition Function, Loss, Estimator, Unbiased, Bias, Variance, Predictive Entropy, Entropy, Mutual Information, Bayesian Neural Networks, BNN, Neural Networks, Deep Ensembles, ResNet}

\vskip 0.3in
]

\printAffiliationsAndNotice{\icmlEqualContribution} %

\begin{abstract}
We introduce a new framework for sample-efficient model evaluation that we call \emph{active testing}.
While approaches like active learning reduce the number of labels needed for model \emph{training}, existing literature largely ignores the cost of labeling \emph{test} data, typically unrealistically assuming large test sets for model evaluation.
This creates a disconnect to real applications, where test labels are important and just as expensive, e.g.~for optimizing hyperparameters.
Active testing addresses this by carefully selecting the test points to label, ensuring model evaluation is sample-efficient.
To this end, we derive theoretically-grounded and intuitive acquisition strategies that are specifically tailored to the goals of active testing, noting these are distinct to those of active learning.
As actively selecting labels introduces a bias; we further show how to remove this bias while reducing the variance of the estimator at the same time.
Active testing is easy to implement and can be applied to any supervised machine learning method.
We demonstrate its effectiveness on models including WideResNets and Gaussian processes on datasets including Fashion-MNIST and CIFAR-100.
\end{abstract}

\section{Introduction}
Although unlabeled datapoints are often plentiful, labels can be expensive.
For example, in scientific applications acquiring a single label can require expert researchers and weeks of lab time.
However, some labels are more informative than others.
In principle, this means that we can pick the most useful points to spend our budget wisely.

These ideas have motivated extensive research into actively selecting \emph{training} labels \citep{atlas_training_1990, Settles2010}, but the cost of labeling \emph{test} data has been largely ignored \citep{lowell_practical_2019}.
In artificial research settings, this is often not a problem: we can `cheat' by using enormous test datasets if the goal is to see how good some sample-efficient training approach is.
But for practitioners this creates a huge issue:
in practice, one must evaluate model performance, both to choose the best model and to develop trust in individual models.
Whenever labels are expensive enough that we need to carefully pick training data, we cannot afford to be wasteful with test data either.

\begin{figure}[!t]
\centering
\includegraphics{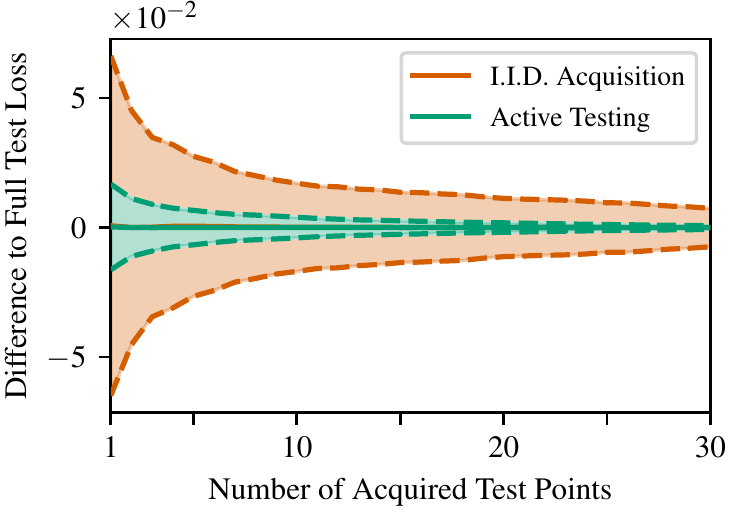}
\vspace{-6pt}
\ifvspace \vspace{-3mm} \fi 
\caption{%
    \textcolor{pal2}{Active testing} estimates the test loss much more precisely than \textcolor{pal3}{uniform sampling} for the same number of labeled test points.
    Active data selection during testing resolves a major barrier to sample-efficient machine learning, complementing prior work which has focused only on training.
    For details, see \S\ref{sec:exp_synthetic}.
    \vspace{-10pt}
    }
\label{fig:intro}
\end{figure}

To address this, we introduce a framework for actively selecting test points for efficient labeling that we call \emph{active testing}.
To this end, we derive acquisition functions with which to select test points to maximize the accuracy of the resulting empirical risk estimate.
We find that the principles that make these acquisition functions effective are quite different from their active learning counterparts.
Given a fixed budget, we can then estimate quantities like the test loss much more accurately than naively labeling points at random.
An example of this is given in~\cref{fig:intro}.

Starting with an idealized, but intractable, approach, we show how practical acquisition strategies for active testing can be derived.
In particular, we derive specific principled acquisition frameworks for classification and regression.
Each of these depends only on a predictive model for outputs, allowing for substantial customization of the framework to particular problems and computational budgets.
We realize the flexibility of this framework by showing how one can implement fast, simple, and surprisingly effective methods that rely only on the original model, as well as much more powerful techniques which use a `surrogate' model that captures information from already acquired test labels and accounts for errors and potential overconfidence in the original model.

A difficulty in active testing is that choosing test data using information from the training data or the model being evaluated creates a sample-selection bias \citep{mackay_information-based_1992, dasgupta_hierarchical_2008}.
For example, acquiring points where the model is least certain~\citep{houlsby_bayesian_2011} will likely overestimate the test loss: the least certain points will tend to be harder than average.
Moreover, the effect will be stronger for overconfident models, undermining our ability to select models or optimize hyperparameters.
We show how to remove this bias using the weighting scheme introduced by \citet{Farquhar2021Statistical}. %
The recency of this technique is perhaps why the extensive literature on active learning has neglected actively selecting test data;
this bias is far more harmful for testing than it is for training.

Our approach is general and applies to practically any machine learning model and task, not just settings where active learning is used.
We show active testing of standard neural networks, Bayesian neural networks, Gaussian processes, and random forests from toy regression data to image classification problems like CIFAR-100.
While our acquisition strategies provide a starting point for the field, we expect there to be considerable room for further innovation in active testing, much like the vast array of approaches developed for active learning.

In summary, our main contributions are:
\begin{itemize}
    \item We formalize active testing as a framework for sample-efficient model evaluation (\S\ref{s:sample_efficient}).
    \item We derive principled acquisition strategies for regression and 
    classification~(\S\ref{s:acquisition_functions}).
    \item We empirically show that our method yields sample-efficient and unbiased model evaluations, significantly reducing the number of labels required for testing~(\S\ref{s:experiments}).
\end{itemize}

\section{Active Testing}\label{s:sample_efficient}
In this section, we introduce the active testing framework.
For now, we set aside the question of how to design the acquisition scheme itself, which is covered in \S\ref{s:acquisition_functions}.
We start with a model which we wish to evaluate, $f: \mathcal{X} \to \mathcal{Y}$, with inputs $\xin \in \mathcal{X}$.
Note that $f$ is \emph{fixed as given}: we will evaluate it, and it will not change during evaluation.
We make very few assumptions about the model.
It could be parametric/non-parametric, stochastic/deterministic and could be applied to any supervised machine learning task.

Our goal is to estimate some model evaluation statistic in a sample-efficient way.
Depending on the setting, this could be a test accuracy, mean squared error, negative log-likelihood, or something else.
For sake of generality, we can write the `test loss' for arbitrary loss functions $\Ls$ evaluated over a test set $\Dt$ of size $N$ as
\begin{equation}
    \hat{R} = \frac{1}{N} \sum_{i_n \in \Dt} \Ls(f(\xin_{i_n}), y_{i_n})\, .
\end{equation}
This test loss is an unbiased estimate for the true risk $R=\E{}{\Ls(f(\xin),y)}$ and is what we would be able to calculate if we possessed labels for every point in the test set.
However, for active testing, we cannot afford to label all the test data.
Instead, we can label only a subset $\Dto \subseteq \Dt$.
Although we could choose all elements of $\Dto$ in a single step, doing so is sub-optimal as information garnered from previously acquired labels can be used to select future points more effectively.
Thus, we will pre-emptively introduce an index $m$ tracking the labeling order:
at each step $m$, we acquire the label $y_{i_m}$ for the point with index $i_m$ and add this point to $\Dto$.

\subsection{A Naive Baseline}
Standard practice ignores the cost of labeling the test set---it does not actively pick the test points.
For a labeling budget $M$, this is equivalent to uniformly sampling a subset of the test data and then calculating the subsample empirical risk
\begin{equation}
    \Riid = \frac{1}{M} \sum_{i_m \in \Dto} \Ls(f(\xin_{i_m}), y_{i_m}). \label{eq:r_iid}
\end{equation}
Uniform sampling guarantees the data are independently and identically distributed (i.i.d.) so that the estimate is 
unbiased, $\mathbb{E} [\Riid]=\hat{R}$, and converges to the
empirical test risk, $\Riid \to \hat{R}$ as $M\to N$.
However, although this estimator is unbiased its \emph{variance} can be high in the typical setting of $M\ll N$.
That is, on any given run the test loss estimated according to this method might be very different from $\hat{R}$, even though they will be equal in expectation.

\subsection{Actively Sampling Test Points}
To improve on this naive baseline, we need to reduce the variance of the estimator.
A key idea of this work is that this can be done by \emph{actively selecting} the most useful test points to label.
Unfortunately, doing this naively will introduce unwanted and highly problematic bias into our estimates.

In the context of pool-based active learning, \citet{Farquhar2021Statistical} showed that biases from active selection can be corrected by using a stochastic acquisition process and formulating an importance sampling estimator.
Namely, they introduce an \emph{acquisition distribution} $q(i_m)$ that denotes the probability of selecting index $i_m$ to be labeled.
They then compute the Monte Carlo estimator $\Rf$ which, in active testing setting, takes the form
\begin{align}
    \Rf &= \frac{1}{M} \sum_{m = 1}^{M} v_m \Ls\left(f(\xin_{i_m}), y_{i_m}\right), \label{eq:cure}
\end{align}
where $M$ is the size of $\Dto$, $N$ is the size of $\Dt$, and
\begin{align}
 v_m &= 1 + \frac{N - M}{N -m}
 \left(\frac{1}{(N - m + 1)q(i_m)} -1\right). \label{eq:cure_weight}
\end{align}
Not only does $\Rf$ correct the bias of active sampling, if the proposal $q(i_m)$ is suitable it can also (drastically) reduce the variance of the resulting risk estimates compared to both $\Riid$ as well as naively applying active sampling without bias correction.
This is because $\Rf$ is based on importance sampling: a technique designed precisely for reducing variance through appropriate proposals
\citep{kahn1953methods,kahn1955use,owen2013monte}.

Importantly, there are no restrictions on how $q(i_m)$ can depend on the data, and in our context $q(i_m)$ is actually shorthand for $q(i_m; i_{1:m-1}, \Dt, \mathcal{D}_{\textup{train}})$.
This means that we will be able use proposals that depend on the already acquired test data, as well as the training data and the trained model, as we explain in the next section.

\section{Acquisition Functions for Active Testing}\label{s:acquisition_functions}
In the last section, we showed how to construct an unbiased estimator of the test risk using actively sampled test data.
This is exactly the quantity that the practitioner cares about for evaluating a model.
For an estimator to be sample-efficient, its variance should be as small as possible for any given number of labels, $M$.
We now use this principle to derive acquisition proposal distributions (i.e.~acquisition functions) for active testing by constructing an idealized proposal and then showing how it can be approximated.

\subsection{General Framework}\label{s:gen_frame}

As shown by~\citet{Farquhar2021Statistical}, the optimal oracle proposal for $\Rf$ is to sample in proportion to the true loss of each data point, resulting in a single-sample zero-variance estimate of the risk.
In practice, we cannot know the true loss before we have access to the actual label.
In particular, the true distribution of outputs for a given input is typically noisy and we can never know this noise without evaluating the label.
In the context of deriving an unbiased Monte Carlo estimator, the best we can ever hope to achieve is to sample from the expected loss over the true $y\mid \xin_{i_m}$,\footnote{For estimators other than $\Rf$, e.g.~ones based on quadrature schemes or direct regression of the loss, this may no longer be true.}
\begin{equation}
    q^*(i_{m}) \propto \E{p(y\mid\xin_{i_m})}{\Ls(f(\xin_{i_m}), y)}. \label{eq:qfirst}
\end{equation}
Note that as $i_m$ can only take on a finite set of values, the required normalization can be performed straightforwardly.

Of course, $q^*(i_{m})$ remains intractable because we do not know the true distribution for $y|\xin_{i_m}$.
We need to approximate it for unlabeled $\xin_{i_m}$ in a way that captures regions where $f(\xin)$ is a poor predictive model as these will contribute the most to the loss.
This can be hard as $f(\xin)$ itself has already been designed to approximate~$y|\xin$.

Thankfully, we have the following tools at our disposal to deal with this:
(a)~We can incorporate \emph{uncertainty} to identify regions with a lack of available information (e.g.~regions far from any of the training data); (b)~We can introduce \emph{diversity} in our predictions compared to $f(\xin)$ (thereby ensuring that mistakes we make are as distinct as possible to those of $f(\xin)$); and~(c) as we label new points in the test set, we can obtain \emph{more accurate} predictions than $f(\xin)$ by incorporating these additional points.
These essential strategies will help us identify regions where $f(\xin)$ provides a poor fit.
We give examples of how we incorporate them in practice in \S\ref{s:surrogate}.

We now introduce a general framework for approximating $q^{*}(i_m)$ that allows us to use these mechanisms as best as possible.
The starting point for this is to consider the concept of a \emph{surrogate} for $y|\xin$, where we introduce some potentially infinite set of parameters~$\theta$, a corresponding generative model $\pi(\theta)\pi(y|\xin,\theta)$, and then approximate the true $p(y|\xin)$ using the marginal distribution $\pi(y|\xin)=\E{\pi(\theta)}{\pi(y|\xin,\theta)}$ of the surrogate.
We can now approximate $q^*(i_m)$ as
\begin{equation}
\label{eq:optimal_target}
    q(i_m) \propto \E{\pi(\theta)\pi(y|\xin_{i_m},\theta)}{\Ls(f(\xin_{i_m}), y)}.
\end{equation}
With $\theta$ we represent our subjective uncertainty over the outcomes in a principled way.
However, our derivations will lead to acquisition strategies also compatible with discriminative, rather than generative, surrogates, for which $\theta$ will be implicit.

\subsection{Illustrative Example} \label{sec:experiments_basic}
\begin{figure}[!t]
\centering
\ifvspace \vspace{-1mm} \fi 
\includegraphics{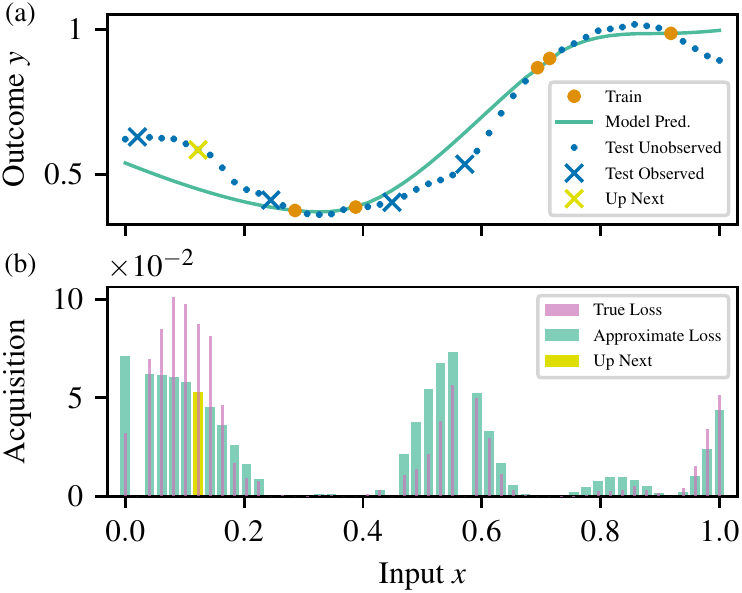}
\ifvspace \vspace{-2mm} \vspace{-0.04in} \fi 
\caption{%
    Illustration of a single active testing step.
    (a) The model has been \textcolor{pal1}{trained} on five points and we currently have observed four \textcolor{pal0}{test} points. (b) We assign acquisition probabilities using the \textcolor{pal2}{estimated} loss of potential test points. Because we do not have access to the true labels, these estimates are different from the \textcolor{pal4}{true} loss. Our \textcolor{pal8}{next} acquisition is then \emph{sampled} from this distribution.
    \vspace{-12pt}
    }
\label{fig:illustrative}
\end{figure}
\Cref{fig:illustrative} shows how active testing chooses the \textcolor{pal8}{next} test point among all available \textcolor{pal0}{test} data.
The model, here a Gaussian process~\citep{rasmussen2003gaussian}, has been trained using the \textcolor{pal1}{training} data and we have already acquired some test points (\textcolor{pal0}{crosses}).
\Cref{fig:illustrative}~(b) shows the \textcolor{pal4}{true} loss known only to an oracle.
Our \textcolor{pal2}{approximate} expected loss is a good proxy in some parts of the input space and worse elsewhere.
The next point is selected by sampling proportionately to the approximate expected loss.
In this example, the surrogate is a Gaussian process that is retrained whenever new labels are observed.
The closer the approximate expected loss is to the true loss, the lower the variance of the estimator $\Rf$ will be;
the estimator will always be unbiased.

\subsection{Deriving Acquisition Functions}\label{s:acquisition_derivations}
We now give principled derivations leading to acquisition functions for a variety of widely-used loss functions.

\textbf{Regression.}
Substituting the \emph{squared error loss} $\Ls(f(\xin), y) = (f(\xin) - y)^2$ into~\eqref{eq:optimal_target} yields
\begin{align}
 q(i_m) &\propto \E{\pi(y\mid\xin_{i_m})}{(f(\xin_{i_m}) - y)^2},
\end{align}
and if we apply a bias-variance decomposition this becomes
\begin{align}
q(i_m) &\propto \underbrace{(f(\xin_{i_m}) - \E{\pi(y\mid \xin_{i_m})}{y})^2}_{\tiny \circled{1}} + \underbrace{\Var{\pi(y\mid\xin_{i_m})}{y}}_{\tiny \circled{2}}. \label{eq:full_expression}   
\end{align}
Here, $\tiny \circled{1}$ is the squared difference between our model prediction $f(\xin_{i_m})$ and the mean prediction of the surrogate: it measures how wrong the surrogate believes the prediction to be.
$\tiny \circled{2}$ is the predictive variance: the uncertainty that the surrogate has about $y$ at $\xin_{i_m}$.

Both $\tiny \circled{1}$ and $\tiny \circled{2}$ are readily accessible in models such as Gaussian processes or Bayesian neural networks.
However, we actually do not need an explicit $\pi(\theta)$ to acquire with~\eqref{eq:full_expression}: we only need to provide approximations for the mean prediction $\E{\pi(y|\xin_{i_m})}{y}$ and predictive variance $\Var{\pi(y|\xin_{i_m})}{y}$. For example, these exist for `deep ensembles' \citep{deep_ensembles} that compute mean and variance predictions from a set of standard neural networks.

A critical subtlety to appreciate is that $\tiny \circled{2}$ incorporates both aleatoric \emph{and} epistemic uncertainty \citep{kendall_what_2017}.
It is not our estimate for the level of noise in $y|\xin_{i_m}$ but the variance of our subjective beliefs for what the value of $y$ \emph{could} be at $\xin_{i_m}$.
This is perhaps easiest to see by noting that 
\begin{align}
\begin{split}
\Var{\pi(y\mid\xin_{i_m})}{y} = & \, %
\Var{\pi(\theta)}{\E{\pi(y\mid\xin_{i_m}, \theta)}{y}} \\
 &+ \E{\pi(\theta)}{\Var{\pi(y\mid\xin_{i_m}, \theta)}{y}}, \label{eq:variance_decomposition}
 \end{split}
\end{align}
where the first term is the variance in our mean prediction and represents our epistemic uncertainty and the latter is our mean prediction of the `aleatoric variance' (label noise).
This is why our construction using $\theta$ is crucial: %
it stresses that \eqref{eq:full_expression} should also take epistemic uncertainty into account.

For regression models with Gaussian outputs $\mathcal{N}(f(\xin), \sigma^2)$, the \emph{negative log-likelihood loss} function and the squared error are related by affine transformation---and following \eqref{eq:optimal_target} so are the acquisition functions.

\textbf{Classification.} %
For classification, predictions $f(\xin)$ generally take the form of conditional probabilities over outcomes $y \in \{1,\dots, C\}$.
First, we study \emph{cross-entropy} %
\begin{align}
    \Ls(f(\xin), y)= - \log f(\xin)_{y}.  \label{eq:ce_loss}
\end{align}
Here, we again introduce a surrogate, %
and, using \eqref{eq:optimal_target}, obtain
\begin{align}
    q(i_m) &\propto \E{\pi(y\mid\xin_{i_m})}{-\log f(\xin_{i_m})_y}.
\end{align}
Now expanding the expectation over $y$ yields
\begin{align}
    q(i_m) &\propto - \sum_{y} \pi(y\mid\xin_{i_m}) \log f(\xin_{i_m})_y, \label{eq:ce_acquisition}
\end{align}
which is the cross-entropy between the marginal predictive distribution of our surrogate, $\pi(y\mid \xin_{i_m})$,
and our model.

We can also derive acquisition strategies based on \emph{accuracy}.
Namely, writing one minus accuracy to obtain a loss,
\begin{align}
    \Ls(f(\xin), y) = 1 - \mathds{1}[y=\arg\max_{y^\prime} f(\xin)_{y^\prime}],
\end{align}
and substituting into~\eqref{eq:optimal_target} yields
\begin{align}
 q(i_m)\propto 1 - \pi(y=y^*(\xin_{i_m})\mid\xin_{i_m}),
\end{align}
where $y^*(\xin_{i_m})=\arg\max_{y^\prime} f(\xin_{i_m})_{y^\prime}$.

\subsection{Tactics for Obtaining Good Surrogates}\label{s:surrogate}
In \S\ref{s:gen_frame} we introduced three ways for the surrogate to assist in finding high-loss regions of $f(\xin)$:
we want it to (a) account for uncertainty over the outcomes, (b) make predictions that are diverse to $f(\xin)$, and (c) incorporate information from all available data.
Motivated by this, we apply the following tactics to obtain good surrogates:

\textbf{Uncertainty.} We should use surrogates that incorporate both epistemic and aleatoric uncertainty effectively,
and further ensure that these are well-calibrated.
Capturing epistemic uncertainty is essential to predicting regions of high loss,
while aleatoric uncertainty still contributes and cannot be ignored, particularly if heteroscedastic.
A variety of different approaches can be effective in this regard and thus provide successful surrogates. For example, Bayesian neural networks, deep ensembles, and Gaussian processes. 

\textbf{Fidelity.} In real-world settings, $f$ may be constrained to be memory-efficient, fast, or interpretable. If labels are expensive enough, we can relax these constraints at test time and construct a more capable surrogate.
In fact, we practically find that using an ensemble of models like $f$ is a robust way of achieving sample-efficiency.

{\setlength{\textfloatsep}{12pt}
\renewcommand{\algorithmicrequire}{\textbf{Input:}}
\begin{algorithm}[t]
    \caption{Active Testing}
    \label{alg:main}
    \begin{algorithmic}[1]
        \Require{Model $f$ trained on data $\Dtr$}
        \State Train surrogate $\pi$ \& choose acquisition proposal form $q$ %
        \For{$m=1$ to $M$}
            \State $i_m \sim q(i_m;\pi)$, observe $y_{i_m}$, add to $\Dto$ 
             \State Calculate $\Ls(f(\xin_{i_m}), y_{i_m})$ and $v_m$ \Comment \cref{eq:cure_weight}
             \State Update $\pi$, e.g.~retraining on $\Dtr\cup\Dto$ %
        \EndFor
        \State Return $\Rf$ \Comment \cref{eq:cure}
    \end{algorithmic}
\end{algorithm}}

\textbf{Diversity.} By choosing the surrogate from a different model family or adjusting its hyperparameters, we can decorrelate the errors of the surrogate and $f$, resulting in better exploration.
For example, we find that random forests~\citep{breiman2001random} can help evaluate neural networks.

\textbf{Extra data.}
If our computational constraints are not critical, we should retrain the surrogate on $\Dto\cup\Dtr$ after each step.
The exposure to additional data will make the surrogate a better approximation of the true outcomes.

\textbf{Thompson-Ensemble.} Retraining the surrogate
can also create diversity in predictions due to stochasticity in the training process, the addition of new data,
or even deliberate randomization.
In fact, we can view retraining the surrogate at regular intervals as implicitly defining an ensemble of surrogates, with the
surrogate used at any given iteration forming a Thompson-sample \citep{thompson1933likelihood} from this ensemble.
This will generally be more powerful and more diverse than a single surrogate, providing further motivation
for retraining and potentially even deliberate variations in surrogates/hyperparameters between iterations.

In \S\ref{s:experiments} we empirically assess the relative importance of these considerations, which depends heavily on the situation. %
For example, the benefit of retraining using the labels acquired at test-time is especially large in very low-data settings, while the benefit of ensembling can be large even when there is more data available.
Putting everything together, Algorithm~\ref{alg:main} provides a summary of our general framework.

If compute is at a premium for acquisitions, a simple alternative heuristic is to use our original model for the surrogate.
This avoids learning a new predictive model, but it suffers because now the surrogate can never disagree with $f$.
Instead, we have to rely entirely on uncertainties for approximating \eqref{eq:optimal_target}: for regression, $\tiny \circled{1}$ in \eqref{eq:full_expression} is zero, and for classification, \eqref{eq:ce_acquisition} reduces to the predictive entropy.
In general, we do not recommend this strategy, \emph{unless} computational constraints are substantial \emph{and} there is reason to believe that the epistemic and aleatoric uncertainties from $f$ represent the true loss well.
If the latter is true, this simplistic approach can perform surprisingly well, although it is always outperformed by more complex strategies.
In particular, 
training a single fixed 
surrogate that is distinct from $f$ will still typically provide noticeable benefits.

\begin{figure}[t]
\begin{tcolorbox}[colback=white,colbacktitle=white,coltitle=black,fonttitle=\bfseries,title=Why are acquisition strategies different for active learning and active testing?, parbox=false]
Researchers have already investigated acquisition functions for active learning, and it would be helpful if we could just apply these here.
However, active testing is a different problem conceptually because we are not trying to use the data to fit a model.
\vspace{1mm}

First, popular approaches for active learning avoid areas with high aleatoric uncertainty while seeking out high epistemic uncertainty. %
This motivates acquisition functions like BALD \citep{houlsby_bayesian_2011} or BatchBALD \citep{kirsch_batchbald}.
For active testing, however, areas of high aleatoric uncertainty can be critical to the estimate.
\vspace{1mm}

Second, as \citet{imberg_optimal_2020} point out, the optimal acquisition scheme for active learning will minimize the expected generalization error at the end of training.
They show how this motivates additional terms beyond what one would get from minimizing the variance of the loss estimator.
\vspace{1mm}

Third, as \citet{Farquhar2021Statistical} show, a biased loss estimator can be helpful during training because it often partially cancels the natural bias of the training loss. %
This is no longer true at test-time, where we want to minimize bias as much as possible.
\end{tcolorbox}
\end{figure}

\section{Related Work}\label{s:related_work}

Efficient use of labels is a major aim in machine learning and it is often important to use
large pools of unlabeled data through unsupervised or semi-supervised methods \citep{chapelle2009semi,erhan2010does,kingma2014semi}.

An even more efficient strategy is to collect only data that is likely to be particularly informative in the first place.
Such approaches are known as optimal or adaptive \emph{experimental
design}~\citep{lindley1956,chaloner1995bayesian,sebastiani2000maximum,foster2020unified,foster2021deep} and are typically
formalized through optimizing the (expected) information gained during an experiment.

Perhaps the best-known instance of adaptive experimental design is \emph{active learning},
wherein the designs to be chosen are the data points for which to acquire labels~\citep{atlas_training_1990, Settles2010,houlsby_bayesian_2011,sener_active_2018}.
This is typically done by optimizing, or sampling from, an acquisition function, with
much discussion in the literature on the form this should take~\citep{imberg_optimal_2020}.

What most of this work neglects is the wasteful acquisition of data for \emph{testing}.
\citet{lowell_practical_2019} acknowledge this and describe it as a major barrier to the adoption of active learning methods in practice.
The potential for `active testing' was raised by~\citet{nguyen_active_2018}, but they focused on the special case of noisily annotated labels that must be vetted and did not acknowledge the substantial bias that their method introduces.
\citet{Farquhar2021Statistical} introduce the variance-reducing unbiased estimator for active sampling which we apply.
However, their focus is mostly on correcting the bias of active \emph{learning}~\citep{bach_active_2006, sugiyama_active_2006, beygelzimer_importance_2009, ganti_upal_2012} and they do not consider appropriate acquisition strategies for active testing.
Note that their theoretical results about the properties of $\Rf$ carry over to active setting.

Other methods like Bayesian Quadrature~\cite{rasmussen2003bayesian,osborne2010bayesian} and kernel herding~\cite{chen2012super} can also sometimes employ active selection of points.
Of particular note,~\citet{osborne2012active,OsborneBayesianModel2019} 
study active learning of model evidence in the context of Bayesian Quadrature.

\citet{bennett2010online,katariya2012active,kumar2018classifier,ji2020active} explore the efficient evaluation of classifiers based on stratification, rather than active selection of individual labels.
Namely, they divide the test pool into strata according to simple metrics such as classifier confidence.
Test data are then acquired by first sampling a stratum and then selecting data \emph{uniformly} within.
Sample-efficiency for these approaches could be improved by performing active testing within the strata.
\citet{sawade2010active} similarly explore active risk estimation through importance sampling, but rely on sampling with replacement which is suboptimal in pool-based settings (see \cref{app:sawade}).
Moreover, like the other aforementioned works, they do not consider the use of surrogates to allow for more effective acquisition strategies.

\begin{figure}[!t]
  \centering
  \includegraphics{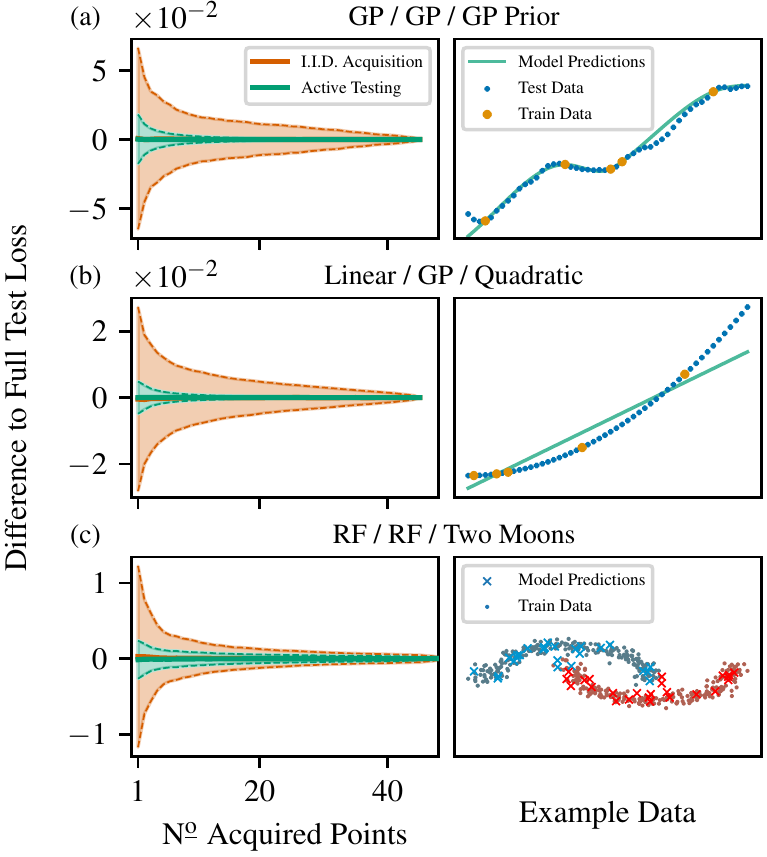}
  \ifvspace \vspace{-2mm} \vspace{-0.05in} \else \vspace{-3mm} \fi
  \caption{%
  Active testing yields unbiased estimates of the test loss with significantly reduced variance.
  Each row shows a different combination of \emph{model/surrogate/data}.
  GP is short for Gaussian process, RF for random forest.
  The first column displays the mean difference of the estimators to the true loss on the full test set (known only to an oracle).
  We retrain surrogates after each acquisition on all observed data.
  Shading indicates standard deviation over \num{5000} (a-b) / \num{2500} (c) runs; data is randomized between runs.
  The second column shows example data with \textcolor{pal2}{model predictions}, and the points used for \textcolor{pal1}{training} and \textcolor{pal0}{testing} (a-b).
  }
    \ifvspace \vspace{-12pt} \fi
  \label{fig:synthetic_comparison}
\end{figure}

\section{Empirical Investigation}\label{s:experiments}
We now assess the empirical performance of active testing and investigate the relative
merits of different strategies.
Similar to active learning, we assume a setting where sample acquisition is expensive, and therefore, per-sample efficiency is critical.
Full details as well as additional results are provided in the appendix, and we release code for reproducing the results at \href{http://github.com/jlko/active-testing}{github.com/jlko/active-testing}.

We note a small but important practicality: we ensure all points have a minimum proposal probability regardless of the acquisition function value, to ensure that the weights are bounded even if $q$ is badly calibrated (cf.~\cref{app:clipping}).

\subsection{Synthetic Data} \label{sec:exp_synthetic}
We first show that active testing on synthetic datasets offers sample-efficient model evaluations.
By way of example, we actively evaluate a Gaussian process~\citep{rasmussen2003bayesian} and a linear model for regression, and a random forest~\citep{breiman2001random} for classification.
For regression, we estimate the squared error and acquire test labels via \cref{eq:full_expression}; for classification, we estimate the cross-entropy loss and acquire with \cref{eq:ce_acquisition}.
We use Gaussian process and random forest surrogates that are retrained on all observed data after each acquisition.

Figure \ref{fig:synthetic_comparison} shows how the difference between our test loss estimation and the truth (known only to an oracle) is much smaller than the naive $\Riid$: active testing allows us to precisely estimate the empirical test loss using far fewer samples. %
For example, after acquiring labels for only \num{5} test points in (a), the standard deviation of active testing is already as low as it is for i.i.d.\ acquisition at step \num{40}, nearly the entire test set.
Further, we can see that the estimates of $\Rf$ are indeed unbiased.
\Cref{app:add_synthetic} gives experiments on additional synthetic datasets.

Here we have actually acquired the full test set.
This lets us show that both $\Rf$ and $\Riid$ converge to the empirical test loss on the entire test set.
However, typically we cannot do this
which makes the difference in variance between $\Riid$ and $\Rf$ at lower acquisition numbers crucial.

\subsection{Surrogate Choice Case Study: Image Classification}
\label{sec:exp_img}
\begin{figure}[t]
  \centering
  \includegraphics{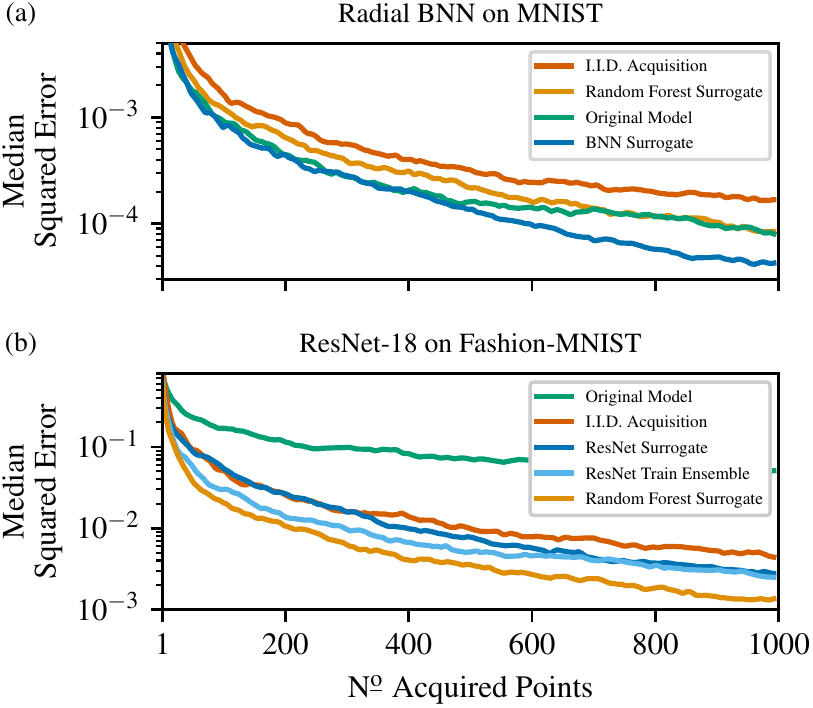}
  \vspace{-12pt}
  \caption{%
    Median squared errors for (a) Radial BNN on \textbf{MNIST} and (b) ResNet-18 on \textbf{Fashion-MNIST} in a small-data setting.
    \textcolor{pal2}{Original Model} samples proportional to predictive entropy, X Surrogate iteratively retrains a surrogate on all observed data, and \textcolor{pal9}{ResNet Train Ensemble} is a deep ensemble trained on $\Dtr$ once.
    Lower is better; medians are over \num{1085} runs for (a), \num{872} for (b).
    \vspace{-10pt} 
  }
  \label{fig:img_small}
\end{figure}
We now investigate the impact of the different surrogate choices. %
For this, we move to more complex image classification tasks and additionally restrict the number of training points to only \num{250}.
This makes it harder to predict the true loss. %
Therefore, the strategies discussed in \S\ref{s:surrogate} are especially important to maximize sample-efficiency.

We evaluate two model types for this examination.
First, a Radial Bayesian Neural Network (Radial BNN) \citep{pmlr-v108-farquhar20a} on the MNIST dataset \citep{lecun1998gradient} in \cref{fig:img_small} (a).
Radial BNNs are a recent approach for variational inference in BNNs \citep{pmlr-v37-blundell15} and we use them because of their well-calibrated uncertainties.
We also evaluate a ResNet\=/18 \citep{resnet} trained on Fashion\=/MNIST \citep{fmnist} in \cref{fig:img_small} (b) to investigate active testing with conventional neural network architectures.
In these figures, we show the median squared error of the different surrogate strategies on a logarithmic scale to highlight differences between the approaches.
While not shown, note that all approaches do still obtain unbiased estimates.
We again use \cref{eq:ce_acquisition} to estimate the cross-entropy loss of the models.

\textbf{Predictive Entropy.} 
We first consider the most naive of the approaches mentioned in \S\ref{s:surrogate}: using the unchanged \textcolor{pal2}{original model} as the surrogate, which leads to acquisitions based on model predictive entropy.
For the Radial BNN, this approach already yields improvements over \textcolor{pal3}{i.i.d.\ acquisition} in \cref{fig:img_small} (a).
The same can not be said for the ResNet in \cref{fig:img_small} (b), for which predictive entropy actually performs worse than i.i.d.\ acquisition.
Presumably, this is the case because the standard neural network struggles to model epistemic uncertainty.
Now, we progress to more complex surrogates, improving performance over the naive approach.

\textbf{Retraining.}
\textcolor{pal0}{BNN surrogate} is a surrogate with identical setup as the original model that is retrained on the total observed data, $\Dtr\cup\Dto$, \num{12} times with increasingly large gaps.
This leads to improved performance over the naive model predictive entropy, especially as more data is acquired.
Similarly, the \textcolor{pal0}{ResNet surrogate} shows much-improved performance over predictive entropy when regularly retrained, now outperforming i.i.d.\ acquisition.

\textbf{Different Model.}
As discussed in \S\ref{s:surrogate}, it may be beneficial to choose the surrogate from a different model family to promote diversity in its predictions.
We use a \textcolor{pal1}{random forest} as a surrogate for both \cref{fig:img_small} (a) and (b).
For the Radial BNN on MNIST, the random forest, while better than i.i.d.\ acquisition, does not improve over the model predictive entropy.
However, for the ResNet on Fashion-MNIST, we find that the random forest surrogate outperforms everything, despite being
a cheaper surrogate. %
This demonstrates that for a surrogate to be successful, it does not necessarily need to be more accurate---although the difference in accuracy is small with so few data.
Instead, the surrogate can be also be successful by being \emph{different} from the original model, i.e.~having structural biases that lead to it making different predictions and therefore discovering mistakes of the original model, with any new mistakes made less important.
Further, if compute is limited, the random forest is attractive because retraining it is much faster.

\textbf{Ensembling Diversity.}
\S\ref{s:surrogate} discussed two ways retraining may help: new data improves the surrogate's predictive model and repeated training promotes diversity through an implicit ensemble.
In \cref{fig:img_small} (b), we introduce the \textcolor{pal9}{ResNet train ensemble}---a deep ensemble of ResNets trained once on $\Dtr$.
This surrogate allows us to isolate the effect of predictive diversity since it is not exposed to any test data through retraining.
We output mean predictions of the ensemble, and find that the deep ensemble can, a little unexpectedly, outperform the \textcolor{pal0}{ResNet surrogate} without accessing the extra data.
This is likely because of better calibrated uncertainties and the increased model capacity.

In summary, we have shown that active testing reduces the number of labeled examples required to get a reliable estimate of the test loss for a Radial BNN on MNIST and ResNet\=/18 on FashionMNIST in a challenging setting,
if appropriate surrogates are chosen.

\subsection{Large-Scale Image Classification}
\begin{figure}[t]
  \centering
  \includegraphics{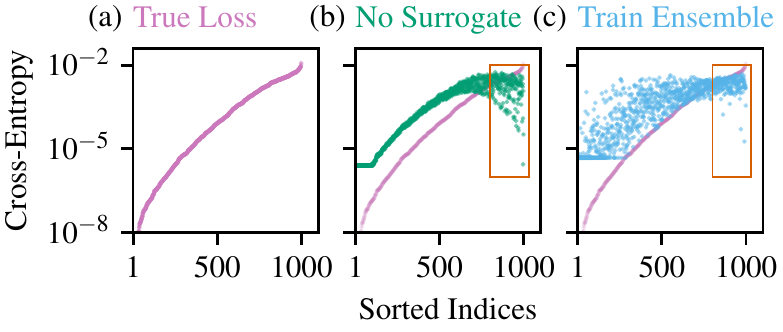}
  \ifvspace \vspace{-2mm} \vspace{-0.06in} \fi
  \caption{%
    \textcolor{pal2}{Predictive entropy} underestimates the \textcolor{pal4}{true loss} of some points by orders of magnitude. Diverse predictions from the \textcolor{pal0}{ensemble of surrogates} help for these crucial high-confidence mistakes, even though they are noisier for low-loss points, improving sample-efficiency overall.
    (a) We sort values of the true losses and use the index order to plot the approximate losses for predictive entropy (b) and an ensemble of surrogates (c), ideally seeing few small approximated losses on the right.
    Shown is a ResNet\=/18 on CIFAR-100; note the log-scale on $y$ and the use of clipping to avoid overly small acquisition probabilities.
    \vspace{-12pt} 
}
  \label{fig:dists}
\end{figure}
\begin{figure}[t]
  \centering
  \includegraphics{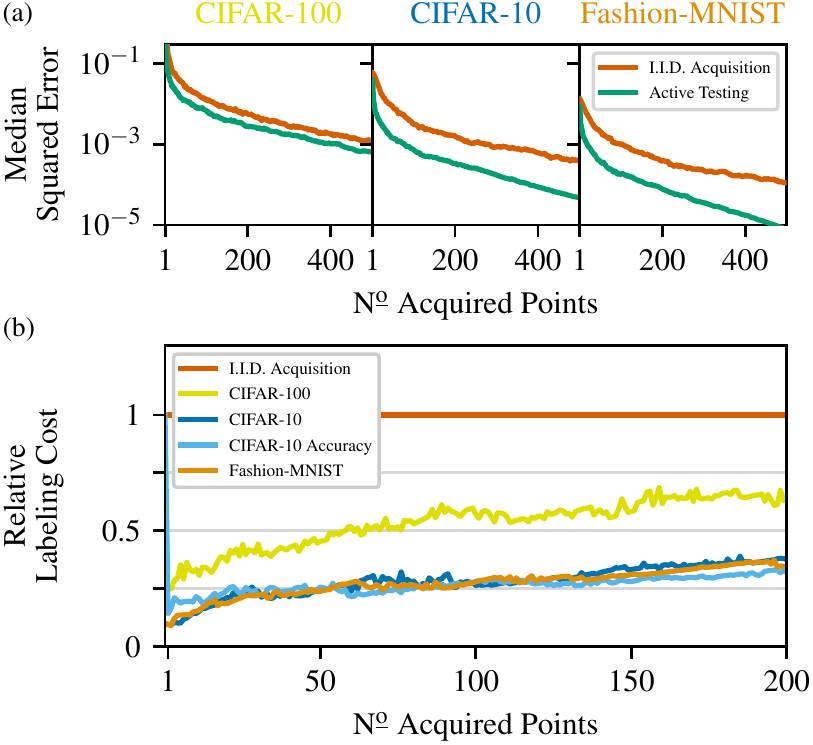}
  \vspace{-12pt}
  \caption{%
    Active testing of a WideResNet on \textbf{\textcolor{pal8}{CIFAR\=/100}} and ResNet\=/18 on \textbf{\textcolor{pal0}{CIFAR\=/10}} and \textbf{\textcolor{pal1}{Fashion\=/MNIST}}.
    (a) Convergences of errors for active testing/i.i.d.~acquisition.
    (b) Relative effective labeling cost. %
    Active testing consistently improves the sample-efficiency.
    Lower is better; medians over \num{1000} random test sets.
  }
  \vspace{-12pt}
  \label{fig:img_large_all}
\end{figure}
We now additionally apply active testing to a Resnet\=/18 trained on CIFAR\=/10 \citep{krizhevsky2009learning} and a WideResNet \citep{BMVC2016_87} trained on CIFAR\=/100.
As the complexity of the datasets increases, it becomes harder to estimate the loss, and hence, it is crucial to show that active testing scales to these scenarios.
We use conventional training set sizes of \num{50000} data points.%

In the previous section, we have seen surrogates based on deep ensembles perform well, even if they are only trained once and not exposed to any acquired test labels.
For the following experiments, we therefore use these ensembles as surrogates.
This is even more justified in the common case where there is much more training data than test data; the extra information in the test labels will typically help less.

In \cref{fig:dists}, we further visualize how ensembles increase the quality of the approximated loss in this setting.
The original model (b) makes overconfident false predictions with high losses which are rarely detected (\textcolor{pal3}{box}).
But the ensemble avoids the majority of these mistakes (c, \textcolor{pal3}{box}) which contribute most to the weighted loss estimator \cref{eq:cure}.

In all cases, the active testing estimator has lower median squared error than the baseline, see \cref{fig:img_large_all} (a)---again note the log-scale.
We further show in \cref{fig:img_large_all} (b) that using active testing is much more sample-efficient than i.i.d.~sampling by calculating the `relative labeling cost': the proportion of actively chosen points needed to get the same performance as naive uniform sampling.
E.g., a cost of \num{0.25} means we need only $\sfrac{1}{4}$ of actively chosen labels to get equivalently precise test loss.
Thus, for the less complex datasets, we see efficiency gains are in the region of a factor of four, while for CIFAR\=/100 they are closer to a factor of two.
We also show that there are similar gains in sample-efficiency when estimating accuracy---`\textcolor{pal9}{CIFAR-10 Accuracy}' in \cref{fig:img_large_all} (b).

\begin{figure}[t]
  \centering
  \includegraphics{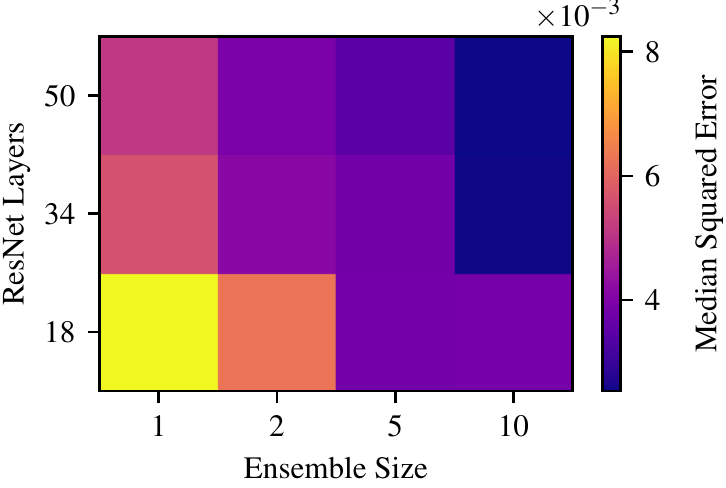}
  \vspace{0mm}
  \caption{%
   Both diversity and fidelity of the surrogate contribute to sample-efficient active testing.
   However, the effect of increasing diversity seems larger than that of increased fidelity.
   We vary the layers (fidelity) and ensemble size (diversity) of the surrogate for active evaluation of a ResNet-18 trained on CIFAR-10.
   Experiments are repeated for \num{1000} randomly drawn test sets and we report average values over acquisition steps \num{100}--\num{200}.
  }
  \label{fig:ablation}
  \vspace{-5mm}
\end{figure}
\begin{figure}[t]
  \centering
  \includegraphics{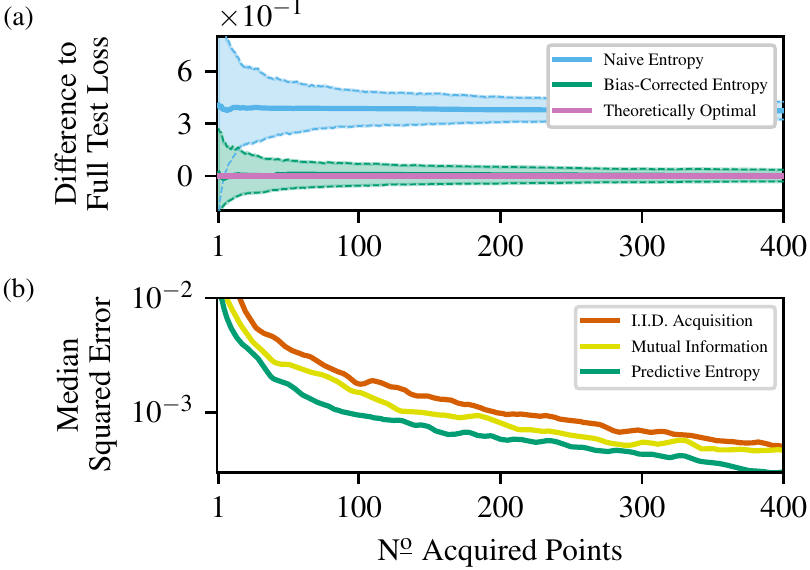}
   \vspace{-5mm}
  \caption{%
    (a) \textcolor{pal9}{Naively} acquiring proportional to the predictive entropy and using the unweighted estimator $\Riid$ leads to biased estimates with high variance compared to \textcolor{pal2}{active testing} with $\Rf$.
    Sampling from the unknown \textcolor{pal4}{true loss} distribution would yield unbiased, zero-variance estimates.
    While this is in practice impossible, the result validates a main theoretical assumption.
    (b)~\textcolor{pal8}{Mutual information}, popular in active learning, underperforms for active testing, even compared to the simple \textcolor{pal2}{predictive entropy} approach. This is because it does not target expected loss.
    Shown for \num{692} runs of a Radial BNN on Fashion-MNIST.
    \vspace{-12pt}
  }
  \label{fig:bias_mi}
\end{figure}
\subsection{Diversity and Fidelity in Ensemble Surrogates}
We now perform an ablation to study the relative effects of surrogate fidelity and diversity on active testing performance.
For this, we evaluate a ResNet-18 trained on CIFAR-10 using different ResNet ensembles as surrogates.
Starting with a base surrogate of a single ResNet-18, we increase the size of the ensemble (mainly increasing diversity) as well as the capacity of the layers (increasing fidelity).
Given the success of the `Train Ensemble' in \S\ref{sec:exp_img}, we train surrogates only once on $\Dtr$, rather than retraining as data is acquired.

As \cref{fig:ablation} shows, both fidelity and diversity contribute to active testing performance:
the best performance is obtained for the most diverse and complex surrogate, justifying our claims in \S\ref{s:surrogate}.
We see that increasing fidelity and diversity both individually help performance, with the effect of the latter seeming to be most pronounced (e.g. an ensemble of 5 Resnet-18s outperforms a single ResNet-50).

\subsection{Optimal Proposals and Unbiasedness}
\cref{fig:bias_mi}~(a) confirms our theoretical assumptions by showing that sampling proportional to the \textcolor{pal4}{true loss}, i.e.\ cheating by asking an oracle for the true outcomes beforehand, does indeed yield exact, single-sample estimates of the loss if combined with $\Rf$.
Further, it confirms the need for a bias-correcting estimator such as $\Rf$: without it, the risk estimates are biased and clearly overestimate model error.

\subsection{Active Testing vs.~Active Learning}
As mentioned in \S\ref{s:surrogate}, we expect there to be differences in acquisition function requirements for active learning and active testing. 
For example, mutual information is a popular acquisition function in active learning \citep{houlsby_bayesian_2011}, but our derivations for classification lead to acquisition strategies based on predictive entropy.
Can mutual information also be used for active testing?
In \cref{fig:bias_mi}~(b) we see that even the simple approach of using the original model as a surrogate and a \textcolor{pal2}{predictive entropy} acquisition outperforms \textcolor{pal8}{mutual information}.
Acquiring with mutual information helps active learning because it focuses on uncertainty that can be reduced by more information rather than irreducible noise.
While this focus helps learning, it is unhelpful for evaluation where all uncertainty is relevant.
This is just one way active testing needs special examination and cannot just re-use results from active learning.

\subsection{Practical Advice}
Empirically, we find that deep learning ensemble surrogates appear to robustly achieve sample-efficient active testing when using our acquisition strategies.
Increases in surrogate fidelity further seem to benefit sample-efficiency.

Active testing generally assumes that acquisitions of labels for samples are expensive, hence we recommend retraining the surrogate whenever new data becomes available.
However, if the cost of this is noticeable relative to that of labeling, our results indicate that not retraining the surrogates is an option, especially when the number of acquired test labels is small compared to the training data.

In general, we do not recommend the naive strategy that relies entirely on the original model and does not introduce a dedicated surrogate model.
As \S\ref{sec:exp_img} has shown, this method can fail to achieve sample-efficient active testing if the original model does not have trustworthy uncertainties.
This strategy should remain a last resort and used only when there is significant reason to trust the original model's uncertainties; we find the diversity provided by a surrogate is critical, even if that surrogate is itself simple.

\section{Conclusions}
\label{sec:disc}
We have introduced the concept of active testing and given principled derivations for acquisition functions suitable for model evaluation.
Active testing allows much more precise estimates of test loss and accuracy using fewer data labels.

While our work provides an exciting starting point for active testing, we believe that the underlying idea of sample-efficient evaluation leaves significant scope for further development and alternative approaches.
We therefore eagerly anticipate what might be achieved with future work.

\section*{Acknowledgements}
We acknowledge funding from the New College Yeotown Scholarship (JK) and Oxford CDT in Cyber Security (SF).

\bibliography{references.bib}
\bibliographystyle{icml2021}
\appendix
\counterwithin{figure}{section}
\newpage
\onecolumn
\ifarxiv 
\icmltitle{Appendix}
\fi 
\FloatBarrier
\section{Additional Experiments}
\label{app:additional_exps}
\subsection{Synthetic Data}
\label{app:add_synthetic}
\begin{figure}[t]
  \centering
  \includegraphics{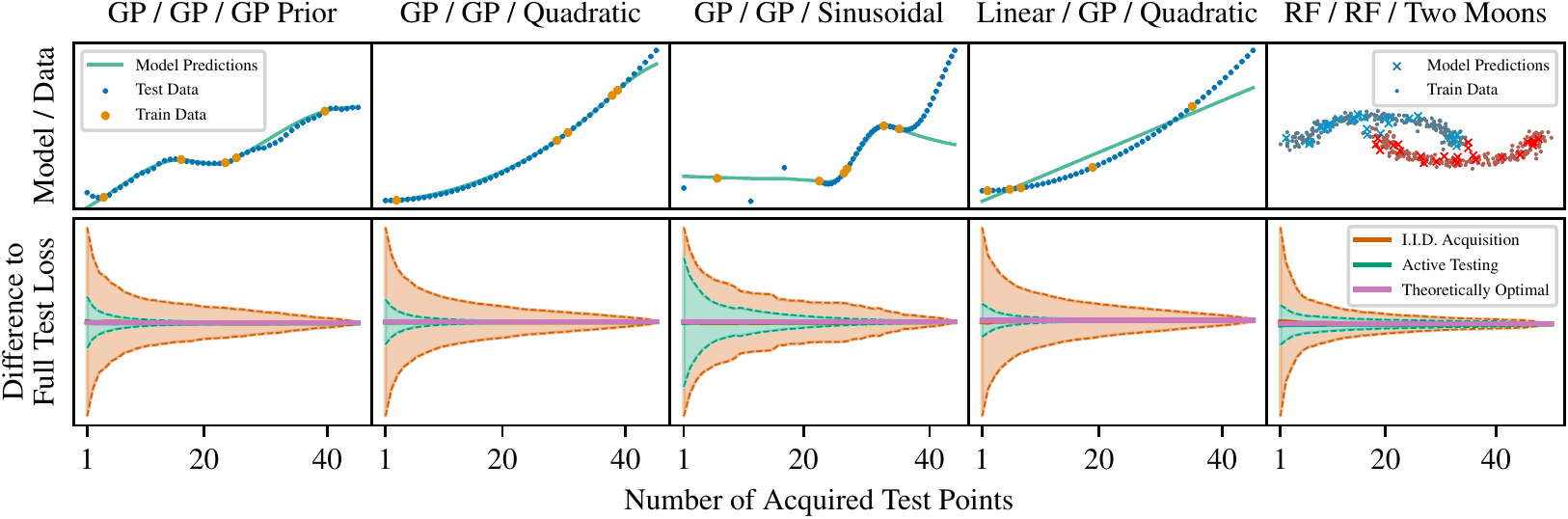}
  \caption{%
  Active testing on synthetic data.
  Columns (1, 4, 5) are repeated from \cref{fig:synthetic_comparison} of the main paper, while (2--3) are exclusive to the appendix.
  Each column shows a different combination of \emph{model/surrogate/data}.
  The first row shows \textcolor{pal2}{model predictions}, as well as the points used for \textcolor{pal1}{training} and  \textcolor{pal0}{testing} for a single draw from the random data generation.
  The second row displays the mean difference of the estimators to the true loss on the full test set (known only to an oracle).
  }
  \label{fig:synthetic_comparison_extended}
\end{figure}

\Cref{fig:synthetic_comparison_extended} shows experiments on additional synthetic data together with the experiments familiar from \cref{fig:synthetic_comparison} of the main paper.
We show two additional regression settings: a GP on sinusoidal data with non-uniform densities in $x$ and a GP on the quadratic data already familiar from \cref{fig:synthetic_comparison}~(b).
Active testing yields gains in sample-efficiency for both of these additional settings.
We give details for all synthetic experiments in \cref{app:synthetic}.

\subsection{Radial BNN on MNIST}
\begin{figure*}[t]
  \centering
  \includegraphics{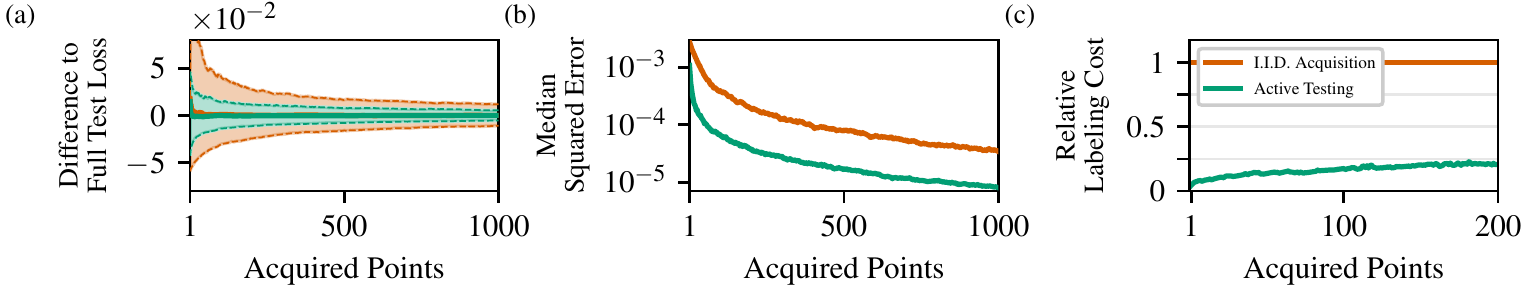}
  \caption{%
     \textcolor{pal2}{Active testing} for a Radial BNN on \textbf{MNIST} using model predictive entropy to estimate test cross-entropy loss.
    (a) Active testing gives unbiased loss estimators with (b) much lower median squared error.
    (c) We calculate the \emph{relative labelling cost}: the sample-efficiency factor of \textcolor{pal3}{i.i.d.~acquisition} to \textcolor{pal2}{active testing} at any number of acquired points.
    Lower is better and values less than \num{1} are gains in sample-efficiency.
    We plot medians and quantiles $(0.1, 0.9)$ and average over \num{992} runs. %
  }
  \label{fig:img_mnist}
\end{figure*}

We also investigate active testing of a Radial BNN on MNIST in a large data regime with \num{50000} training points.
As acquisition strategy, we here simply use the predictive entropy of the Radial BNN, which we observe to be well-calibrated in this simple setting.
Consequently, we see large gains in sample-efficiency of active testing over i.i.d.\ acquisition for this setting as shown in \cref{fig:img_mnist}.
We discuss experimental details for this plot in \cref{app:radial_mnist}

\subsection{Uncertainties and Statistical Significance}
\begin{figure*}[!t]
  \centering
  \includegraphics{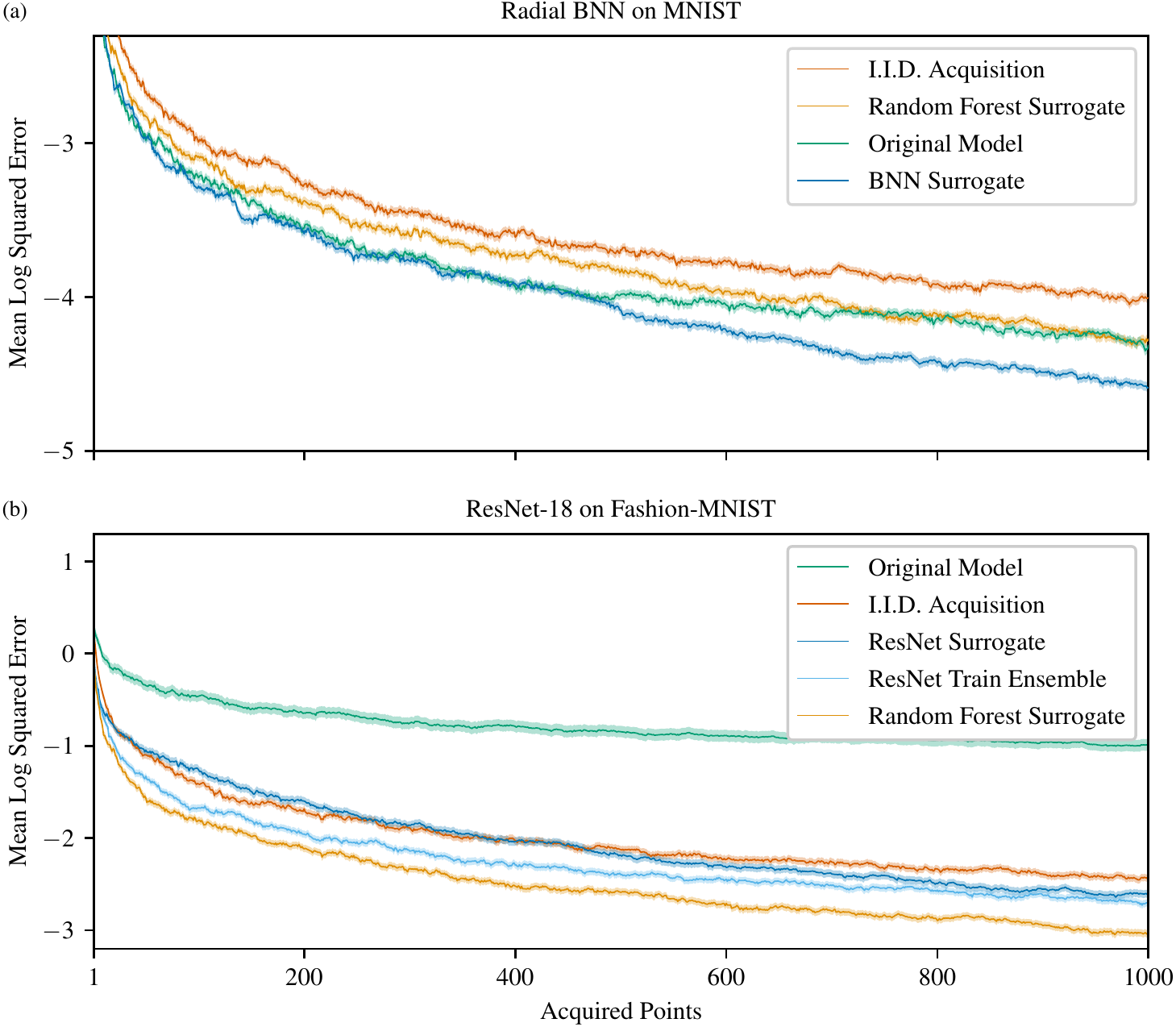}
  \caption{%
    \Cref{fig:img_small} from the main paper but now showing the mean of the log squared difference instead of the median.
    Additionally, shading indicates the standard error of the log squared difference.
    Averages over \num{1085} runs for (a), \num{872} for (b).
    See text and main paper for details.
  }
  \label{fig:A3}
\end{figure*}

\begin{figure*}[!t]
  \centering
  \includegraphics{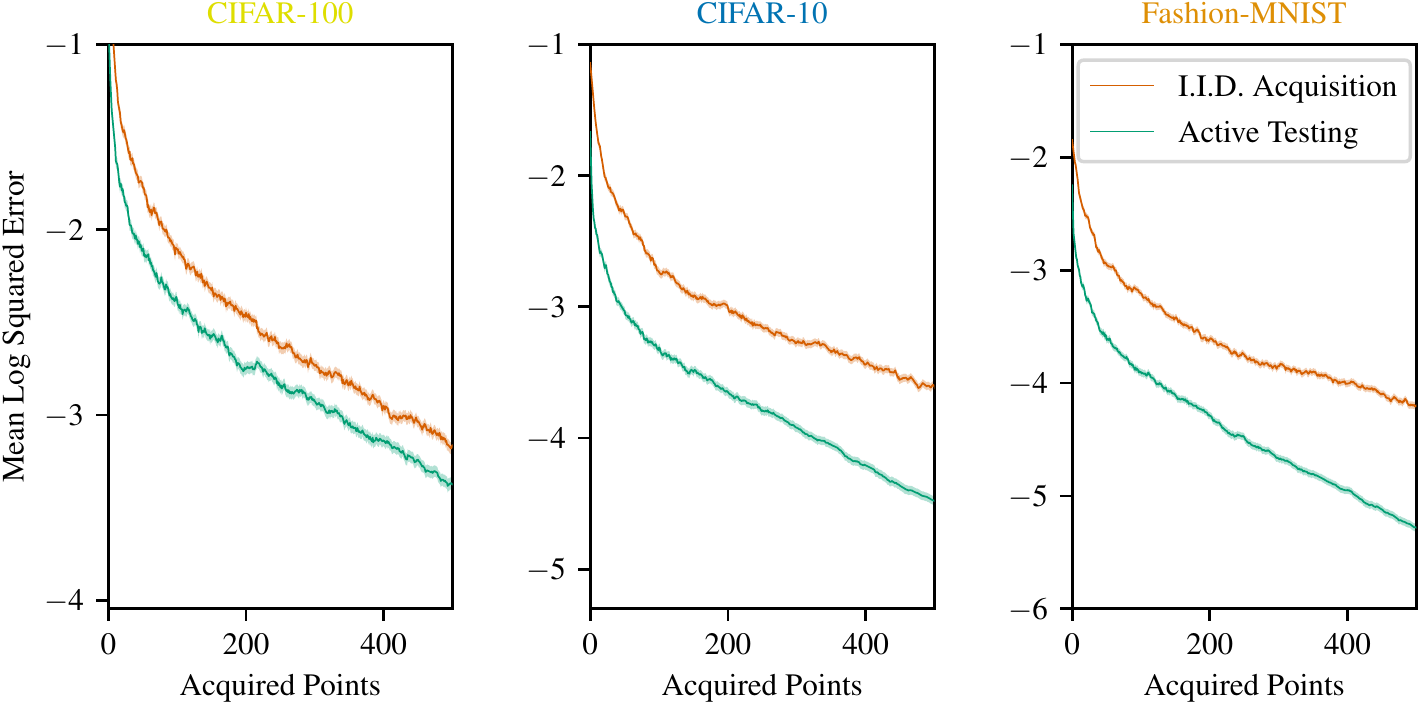}
  \caption{%
    \Cref{fig:img_large_all}~(a) from the main paper but now showing the mean of the log squared difference instead of the median.
    Additionally, shading indicates the standard error of the log squared difference.
    Averages over \num{1000} random test set draws.
    See text and main paper for details.
}
  \label{fig:A4}
\end{figure*}

\begin{figure*}[!t]
  \centering
  \includegraphics{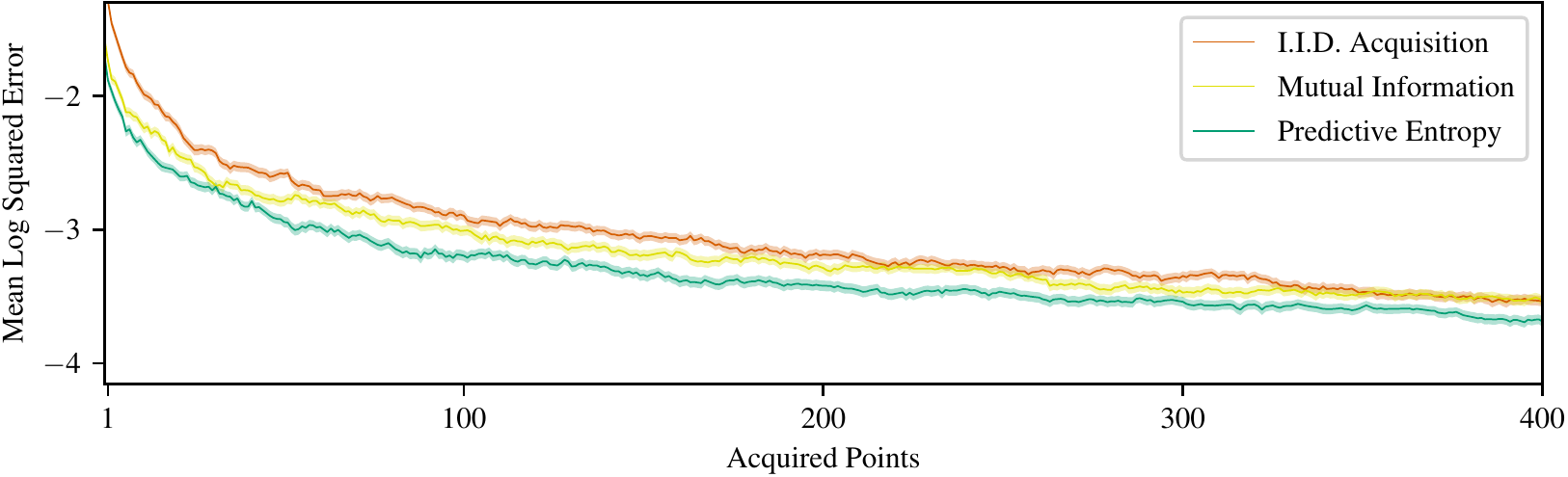}
  \caption{%
    \Cref{fig:bias_mi}~(b) from the main paper but now showing the mean of the log squared difference instead of the median.
    Additionally, shading indicates the standard error of the log squared difference.
    Averages over \num{692} runs.
    See text and main paper for details.
  }
  \label{fig:A5}
\end{figure*}

In \cref{fig:A3,fig:A4,fig:A5}, we show a variation of Figs.~\ref{fig:img_small}, \ref{fig:img_large_all} (a), and \ref{fig:bias_mi} (b) wherein we plot \emph{mean} log squared differences instead of \emph{median} squared differences.\footnote{Reminder: We calculate the squared difference between the true test loss on the full test set (known only to an oracle) and the estimator at each acquisition step.}
All conclusions from the main body of the paper continue to hold for the mean-based visualization.
Additionally, we quantify uncertainties on the mean values via the standard errors of the log squared differences.

We also investigate the statistical significance of the results reported in the paper, computing Wilcoxon signed-rank test statistics to examine if the best active testing strategy has lower population mean rank than all other shown methods.
More precisely, when comparing two methods at a fixed acquisition step, we obtain a pair of samples of squared differences for each run, and we test against the alternative hypothesis that the best performing method has \emph{lower} squared error.
We compare methods at the last displayed acquisition step.
The results of the test show that we can always reject the null hypothesis at \num{5e-3} confidence level.
We give the full results of the Wilcoxon signed-rank tests in \cref{tab:ks}.

\begin{table*}
\caption{Wilcoxon signed-rank tests for experiments from the main paper, comparing the squared differences of the best method against all other shown approaches. The alternative hypothesis is that the `best method' has lower squared error than the `other method'. \\}
\label{tab:ks}
\centering
\begin{tabular}{ r l l l r  l }
\toprule
 Figure & Dataset & Best Method & Other Method & Acq. Step & p-Value \\
 \midrule
 & & & & &  \\[-.8em]
 4 a & MNIST & BNN Surrogate & I.I.D.\ Acquisition & 1000 &  \num{5.438e-61} \\  
  & MNIST & BNN Surrogate & Random Forrest Surrogate & 1000 &\num{7.069e-19 } \\  
 & MNIST & BNN Surrogate & Original Model & 1000 &  \num{9.524e-20} \\  
 4 b & Fashion-MNIST & Random Forrest Surrogate & Original Model &  1000 &\num{6.586e-125} \\  
 & Fashion-MNIST & Random Forrest Surrogate & I.I.D.\ Acquisition &  1000 &\num{4.176e-31} \\
 & Fashion-MNIST & Random Forrest Surrogate & ResNet Surrogate & 1000 & \num{4.249e-17} \\
 & Fashion-MNIST & Random Forrest Surrogate & ResNet Train Ensemble & 1000 & \num{4.254e-10} \\
 6 a & CIFAR-100 & ResNet Train Ensemble & I.I.D.\ Acquisition & 500 &\num{3.010e-11} \\
 & CIFAR-10 & ResNet Train Ensemble & I.I.D.\ Acquisition & 500 & \num{3.988e-76} \\
 & Fashion-MNIST & ResNet Train Ensemble & I.I.D.\ Acquisition & 500 & \num{2.457e-105} \\
 7 & Fashion-MNIST & Predictive Entropy & I.I.D.\ Acquisition & 400 & \num{2.801e-06} \\
 & Fashion-MNIST & Predictive Entropy & Mutual Information & 400 & \num{8.779e-04} \\ \bottomrule
\end{tabular}
\end{table*}

\section{Details on Active Testing}
\subsection{Further Details on Clipping}
\label{app:clipping}
As mentioned briefly in the main paper, we are clipping the predicted $q(i_m)$ to avoid overly small acquisition values.
We do this because $\Rf$ is only unbiased if we have non-zero acquisition probability on \emph{all} points remaining \citep{Farquhar2021Statistical}.
In practice, we bound the value of the acquisition function from below at $\alpha$ times the acquisition probability assigned by a uniform acquisition strategy, where $\alpha=0.2$.

\subsection{Computational Complexity}
\label{app:computational_complexity}
Two components of active testing contribute significantly to computational complexity: Evaluating the acquisition function on the remaining samples of the test pool and retraining the surrogate on all observed samples.

\textbf{Acquisition Function Evaluation.} At each active testing iteration, we require the evaluation of the acquisition function $q$ on all remaining samples in the test pool $\Dt$.
At first iteration, $N$ samples remain in the test pool, contributing $\mathcal{O}(N)$ to the computational complexity.
At each subsequent iteration $m$, $\mathcal{O}(N-m)$ additional evaluations are required.
For a total of $M$ acquisitions, we can bound the cost of evaluating the acquisition function with $\mathcal{O}(MN)$.

\textbf{Surrogate Retraining.}
In our experiments in \S\ref{sec:exp_img}, we demonstrate successful active testing when retraining the surrogate every $K$ steps or training the surrogate only once on $\Dtr$.
However, in active testing practice, label cost may be significantly larger than the cost of retraining.
Therefore, we here assume the worst case $K=M$, i.e.~retraining the surrogate after each acquisition.
This gives maximum sample-efficiency because additional information from newly acquired test labels is directly used to improve surrogate predictions.
At first iteration, only $\Dtr$ is observed and hence retraining costs will be $\mathcal{O}(\lvert \Dtr \rvert)$.
As testing progresses, $m$ samples of the test set become available such that retraining costs grow to $\mathcal{O}(\lvert \Dtr \rvert + m)$ per iteration.
For a total of $M$ acquisitions, retraining costs therefore scale as $\mathcal{O}(M \lvert \Dtr \rvert~+~M^2)$.

Therefore, we can bound the total computational complexity of active testing as $\mathcal{O}(M \lvert \Dtr \rvert + M^2 + MN)$, which can be simplified to $\mathcal{O}(M \lvert \Dtr \rvert + MN)$ as $M\le N$ always.

\section{Details on Experiments}
\label{app:details_on_experiments}

\subsection{Hardware}
For the neural network architectures, we use PyTorch \citep{pytorch}.
We use a mix of NVIDIA RTX 2080, Titan RTX, and K80 GPUs to run the experiments.
No experiment requires more than one GPU in parallel.
Experiments that do not require GPUs, such as the synthetic data experiments, can be run on CPUs only on conventional hardware.

\subsection{\Cref{fig:intro,fig:illustrative,fig:synthetic_comparison}: Synthetic Data}\label{app:synthetic}
We now give details for \cref{fig:synthetic_comparison} of the main paper, shown again in \cref{fig:synthetic_comparison_extended} (columns 1, 4, 5), as well as the additional experiments (columns 3 and 4).
\Cref{fig:intro,fig:illustrative} use the setup of \cref{fig:synthetic_comparison_extended} (column 1).

\textbf{Regression.}
For the Gaussian process data (column 1), we sample $N=50$ points from a Gaussian process with Matern-Kernel $\nu=3/2$ and $l=1$, using the implementation of \citet{sklearn_api}.
For each of the experiments, we sample a different set of points from the Gaussian process prior.
The quadratic data (columns 2, 4) are drawn from $f(x)=y^2$.
The sinusoidal data (column 3) are drawn from $f(x) = \sin(10x) + x^3$, where the density of the samples in $x$ is non-uniform.

\si{5} points are randomly assigned to the training set, \si{45} are used for testing.
The test-train assignments are randomly redrawn between experiments.
For each experiment, we actively evaluate the data with all acquisition functions.
We perform a series of \si{5000} independent experiments.
All regression experiments use a Gaussian process surrogate that is retrained after each acquisition on all observed data and has Matern-Kernel $\nu=3/2$ and $l=1$.

\textbf{Two Moons Classification.}
For the two moons dataset (column 5), we randomly sample $N=500$ points with noise level of \num{0.1} for each experiment, using the implementation of \citet{sklearn_api}.
\si{50} points are randomly assigned to the training set, \si{450} are used for testing.
We perform a series of \si{2500} independent experiments.
We use default parameters and implementation of \citet{sklearn_api} for the random forest.
The surrogate model uses an identical random forest that is retrained after each acquisition on all observed data.

\subsection{\Cref{fig:img_small}: Radial BNN/ResNet-18 on MNIST/Fashion-MNIST}

\textbf{Data.} %
For each experiment, we sample \num{250} training and \num{5000} test points randomly from the combined training and test set of the original datasets.
We standardize the dataset using per-channel mean and standard deviation values over all pixels of the training set.
We actively acquire \num{1000} samples from the test set and compute the `full test loss' on all \num{5000} test points.
We perform stratified sampling for both the train/test as well as the train/val splits.
We retrain the surrogates after acquiring $(0, 5, 10, 20, 30, 40, 100,  250, 400, 550, 700, 850, 1000)$ test points.

\textbf{Radial BNN on MNIST.}
We use the code obtained from \citet{Farquhar2021Statistical} to implement the Radial BNNs.
We use the following hyperparameters, which are default values taken from \citet{Farquhar2021Statistical}:
we use a learning rate of \num{1e-4} and weight decay of \num{1e-4} with the ADAM optimizer \citep{kingma2014adam}, batch size of \num{64}, \num{8} variational samples during training for the BNN, \num{100} variational samples during testing, and convolutions with \num{16} channels.
We train for a maximum of \num{500} epochs with early stopping patience of \num{5} and use validation sets of size \num{50}.
We have not tuned these hyperparameters.

\textbf{ResNet-18 on Fashion-MNIST.}
For Resnet-18, we use the default hyperparameter values introduced by \citet{devries2017improved}:
we use a learning rate of \num{0.1}, weight decay of \num{5e-4}, and momentum of \num{0.9} with an SGD optimizer, and batch size of \num{128}.
We use a cosine annealing schedule for the learning rate as provided by PyTorch.
We train for a maximum of \num{160} epochs with early stopping patience of \num{20} and use validation sets of size \num{50}.
We have not tuned these hyperparameters.

\textbf{Random Forest Surrogate.}
We use random forests with \si{100} estimators, split criterion `entropy', and maximum number of features considered at each split proportional to the square root, otherwise using the default parameters and implementation of \citet{sklearn_api}.
Hyperparameters were set with a single grid-search cross-validation on one particular training set.

\textbf{Training Convergence: Radial BNN on MNIST.}
Training of the BNN takes about \numrange{1}{2} minutes and the early stop patience usually terminates training after around \numrange{50}{100} epochs.
A single experiment run for the Radial BNN, requiring the training of the original model and retraining of all shown surrogates, takes about \num{30} minutes.
The validation accuracy on the initial \num{250} points is at about \numrange{80}{90} percent and grows to about \numrange{90}{98} percent after observing a total of \num{1250} points.
We perform a total of \num{1085} runs.

\textbf{Training Convergence: ResNet-18 on Fashion-MNIST.}
Training of the ResNet takes about \numrange{10}{40} seconds and the early stop patience sets in around \numrange{30}{80} epochs.
A single run for the ResNet, requiring the training of the original model and retraining of all shown surrogates, takes about \num{10} minutes.
The training validation accuracy on the initial \num{250} points is at about \numrange{68}{78} percent and grows to about \numrange{80}{86} percent after observing a total of \num{1250} points.
We perform a total of \num{872} runs.

\subsection{\Cref{fig:dists}: ResNet-18 on CIFAR-100}
The setup for \cref{fig:dists} is a ResNet-18 on CIFAR-100 identical to the experiments from \cref{fig:img_large_all} described in the following.
The model converges to \numrange{72}{76} percent accuracy in about \num{12} minutes and we perform \num{692} runs.

\subsection{\Cref{fig:img_large_all}: ResNet-18/WideResNet on Fashion-MNIST, CIFAR-10/Cifar-100}

\textbf{Data.}
We now respect the original train and test indices of the dataset. We train a model on the training set and then, for each experiment, perform active testing on a subset of \num{1000} randomly sampled points from the original test set of size \num{10000}.
We standardize the dataset using per-channel mean and standard deviation values over all pixels of the training set.
For CIFAR-10 and CIFAR-100, we apply random crops and horizontal flips to the training data in each epoch.
Given the larger training data, we now use validation sets of size \num{2560}.
We re-sample test sets for the next experiment, but keep training sets and models constant.
We perform stratified sampling for both the train/test as well as the train/val splits.
This, we repeat a total of \num{1000} times.

\textbf{ResNet-18.}
Unless otherwise mentioned we use identical hyperparameters as for \cref{fig:img_small}.
On all datasets, we train for a maximum of \num{30} epochs with patience of \num{5}.
Again, we do not systematically tune hyperparameters and only make adjustments to accommodate the increased data size compared to \cref{fig:img_small}.
The model converges to \numrange{93}{94} percent accuracy on Fashion-MNIST in about \num{7} minutes and \numrange{92}{93} percent accuracy on CIFAR-10 in about \num{12} minutes.

\textbf{WideResNet.}
We use a WideResNet of depth \num{40} and the setup introduced by \citet{devries2017improved}:
we train for \num{200} epochs, use a learning rate of \num{0.1}, weight decay of \num{5e-4}, and momentum of \num{0.9} with an SGD optimizer, and batch size of \num{128}.
We also use the scheduler of \citet{devries2017improved} and decrease the learning rate by a factor of $\gamma=0.2$ at epochs \num{60}, \num{120}, and \num{160}.
The model converges to \numrange{78}{80} percent accuracy on CIFAR-100 in about \num{240} minutes.

\textbf{Ensembles.}
For the deep ensemble of Resnet-18s, we use \num{10} models for Fashion-MNIST, \num{5} for CIFAR-10, and \num{15} for CIFAR-100.
For the deep ensemble of WideResnets on CIFAR-100, we use a set of \num{10} models.
The models for the ensembles are identical to the main models, i.e.\ use the same hyperparameters, optimizers, and training setup, and only differ in terms of the optima reached from stochastic model initialization and optimization.

Increasing the size of the ensemble did sometimes yield improved results.
Following initial experiments, we increased the size of the ensemble for Fashion-MNIST from \num{5} to \num{10}.
However, using ensemble size of \num{15} on CIFAR-10 did not improve the results noticeably (nor worsen them), so we display the smaller, and computationally cheaper ensemble in the main body.

\subsection{\Cref{fig:ablation}}
\Cref{fig:ablation} uses the same setup as \cref{fig:img_large_all} for CIFAR-10 with main ResNet-18 model and `train ensemble'-style surrogates.
\FloatBarrier

\subsection{\Cref{fig:bias_mi}}
For the Radial BNN, \cref{fig:bias_mi} uses identical experimental setup as \cref{fig:img_small}, except that the model is trained for a maximum of \num{30} epochs (patience \num{5}) and we use validation sets of size \num{1280}, to account for the increase in data.
For the Fashion-MNIST data, we concatenate the original train and test datasets, from which we then sample \num{50000} points without replacement for the training dataset and \num{10000} for the test dataset.
We acquire a total of \num{1000} points from the test dataset but compute the `full test set loss' on all \num{10000} points.
Splits are stratified by class labels.
We redraw train and test splits, and retrain models between runs.
For each experiment, the model reaches about \numrange{88}{92} percent validation accuracy in about \num{8} minutes.
We perform a total of \num{962} runs.

\subsection{Figure A2: Radial BNN on MNIST.}\label{app:radial_mnist}
For the Radial BNN, \cref{fig:img_mnist} uses identical experimental setup as \cref{fig:img_small}~(a), except that the model is trained for a maximum of \num{50} epochs (patience \num{5}) and we use validation sets of size \num{1280} to account for the increase in data.
For the MNIST data, the setup is identical to \cref{fig:bias_mi}.
The model reaches about \num{98} percent validation accuracy in about \numrange{7}{10} minutes.
We perform a total of \num{992} runs.

\section{Comparison to Active Risk Estimation by Sawade et al.}
\label{app:sawade}
\begin{figure}[t]
    \centering
    \includegraphics{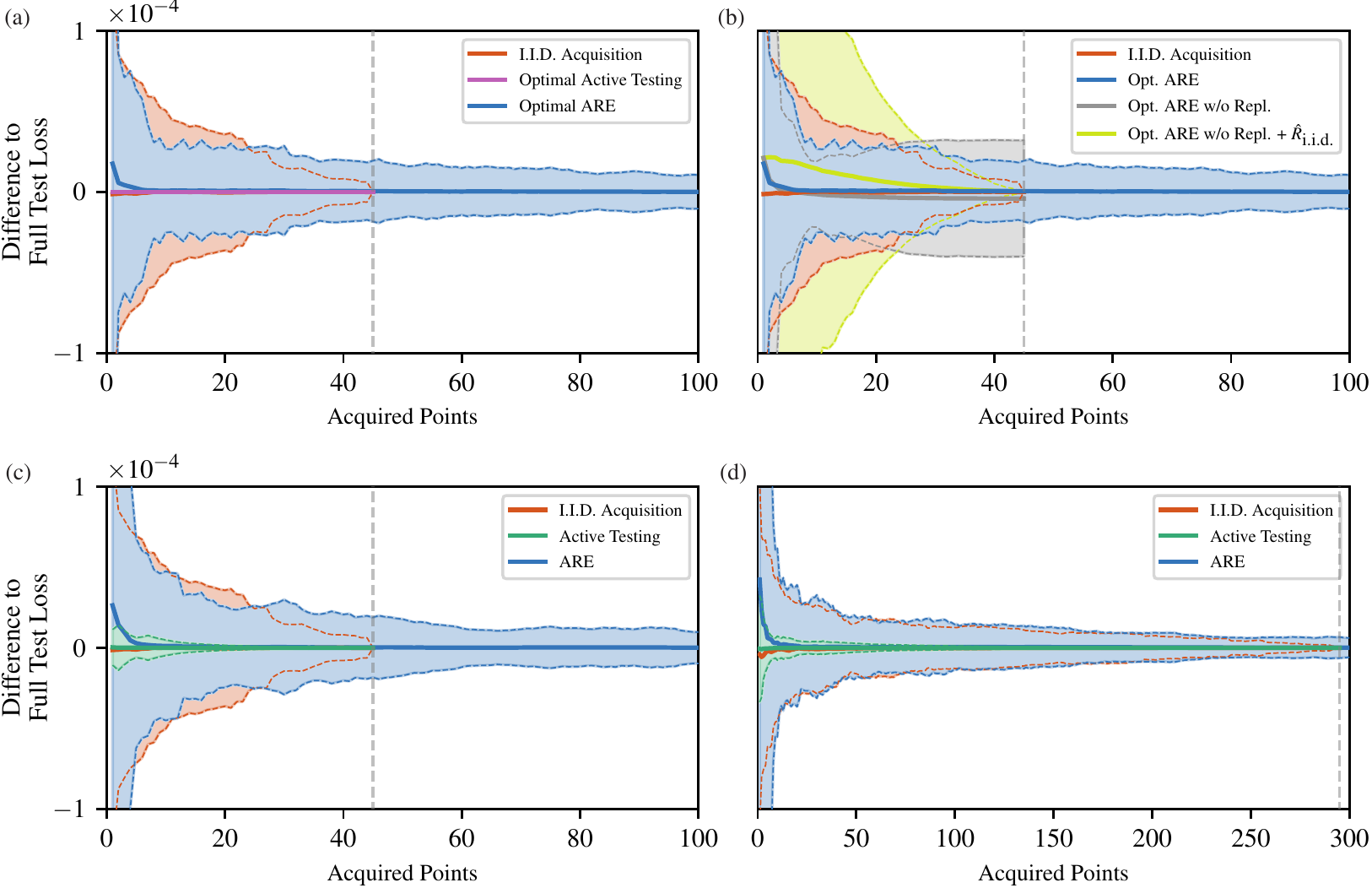}
    \caption{%
        Comparison of Active Testing to Active Risk Estimation (ARE) by \citet{sawade2010active}.
        (a)~While active testing can yield single-sample zero-variance optimal estimates, ARE cannot because it does not fully accommodate the pool-based setting.
        (b)~Naive extensions of ARE to sampling without replacement are unsuccessful.
        (c)~In practice, ARE yields lower performance than active testing.
        (d)~As test set grows, the gap between ARE and active testing will get smaller as sampling with replacement matters less.
        Dashed grey line demarcates number of samples in the test pool.
        Solid lines show means, dashed lines standard deviations over \num{7000} (a--c) / \num{2000} (d) runs.
        }
    \label{fig:sawade}
\end{figure}
We here provide a brief introduction to `Active Risk Estimation' (ARE) by \citet{sawade2010active} as well as an empirical comparison of their approach to ours.
Similar to us, \citet{sawade2010active} actively acquire labels from a test pool of unlabeled samples, aiming to obtain a sample-efficient estimate of the empirical test risk.
However, unlike us, they do not fully accommodate the pool-based setting.
Additionally, \citet{sawade2010active} rely only on the original model to approximate outcomes $y\mid\xin$, while we have shown that appropriate surrogate models can be crucial for sample-efficient active testing.
Moreover, as we show below, active testing outperforms ARE in practice.

\subsection{Background}
\citet{sawade2010active} derive an `optimal acquisition function'
\begin{align}
    q^*(\xin) \propto p(\xin) \sqrt{\int \left[\Ls(f_\theta(\xin), y) - R\right]^2 p(y\mid \xin) \, \mathrm{d}y}, \label{eq:are_optimal}
\end{align}
where $R=\E{}{\Ls(f(\xin),y)}$ is the true model risk.
For the pool-based setting, they obtain an equivalent `optimal acquisition function' by setting $p(\xin_{i_m}) = \frac{1}{M}$ and approximating $R$ with the empirical test risk $\Riid$ over the entire test pool.
As $\Riid$ is still unknown, \citet{sawade2010active} approximate $\Riid$ using the original model's mean predictions $f_\theta(\xin)$ to approximate the true outcomes, giving
\begin{align}
    \hat{R}_{\textup{i.i.d.},\theta} &= \frac{1}{M} \underset{\xin \in \Dt}{\sum} \int \Ls(f_\theta(\xin), \yin) p(\yin|\xin;\theta)\, \mathrm{d}\yin \\
    q^*(\xin) &\propto \sqrt{\int \left[\Ls(f_\theta(\xin), y) - \hat{R}_{\textup{i.i.d.},\theta}\right]^2 p(y\mid \xin;\theta) \, \mathrm{d}y}. \label{eq:are_optimal_approx}
\end{align}
They do not consider the concept of a surrogate model.

Plugging mean squared error loss and Gaussian likelihoods into \eqref{eq:are_optimal_approx}, they derive the acquisition function
\begin{align}
q(\xin) \propto \sqrt{3\sigma_\xin^4 - 2 \hat{R}_{\text{i.i.d.},\theta} \sigma_\xin^2 + \hat{R}_{\text{i.i.d.},\theta}^2}, \label{eq:are_mse}
\end{align}
where $\sigma_\xin^2$ is the total predictive variance (aleatoric and epistemic uncertainty) of the original model's prediction at~$\xin$.

For risk estimation, ARE relies on a standard importance sampling estimator
\begin{align}
    \hat{R}_\textup{IS} = \frac{1}{\overset{M}{\underset{{m=1}}{\sum}} q(\xin_{i_m})} \overset{M}{\underset{{m=1}}{\sum}} \frac{1}{q(\xin_{i_m})} \Ls(f_\theta(\xin_{i_m}, y_{i_m})),
\end{align}
where the uniform $p(\xin_{i_m})$ cancels from the weights.
Unlike $\Rf$, $\hat{R}_\textup{IS}$ requires sampling \emph{with replacement} to obtain unbiased estimates.
In other words, ARE does not remove samples from the test pool after querying them, and often, test points will get sampled repeatedly.

\subsection{Empirical Comparison.}
We compare both theoretically optimal behavior as well as observed practical performance of our active testing and ARE.
For this, we take the familiar setup from \cref{fig:synthetic_comparison_extended}~(column 1).

\textbf{Optimal Behavior.}
We have seen that active testing can ideally lead to single-sample, zero variance estimates of the test loss (cf.~\cref{fig:bias_mi}~(a)).
What is the optimal behavior ARE can obtain?
To test this, we apply \eqref{eq:are_optimal} with the oracle empirical test risk $\Riid$ and the true $y\mid\xin$ (which is a $\delta$-distribution since we generate data without noise).
This constitutes the best case scenario for ARE performance.
\begin{figure}[t]
    \centering
    \includegraphics{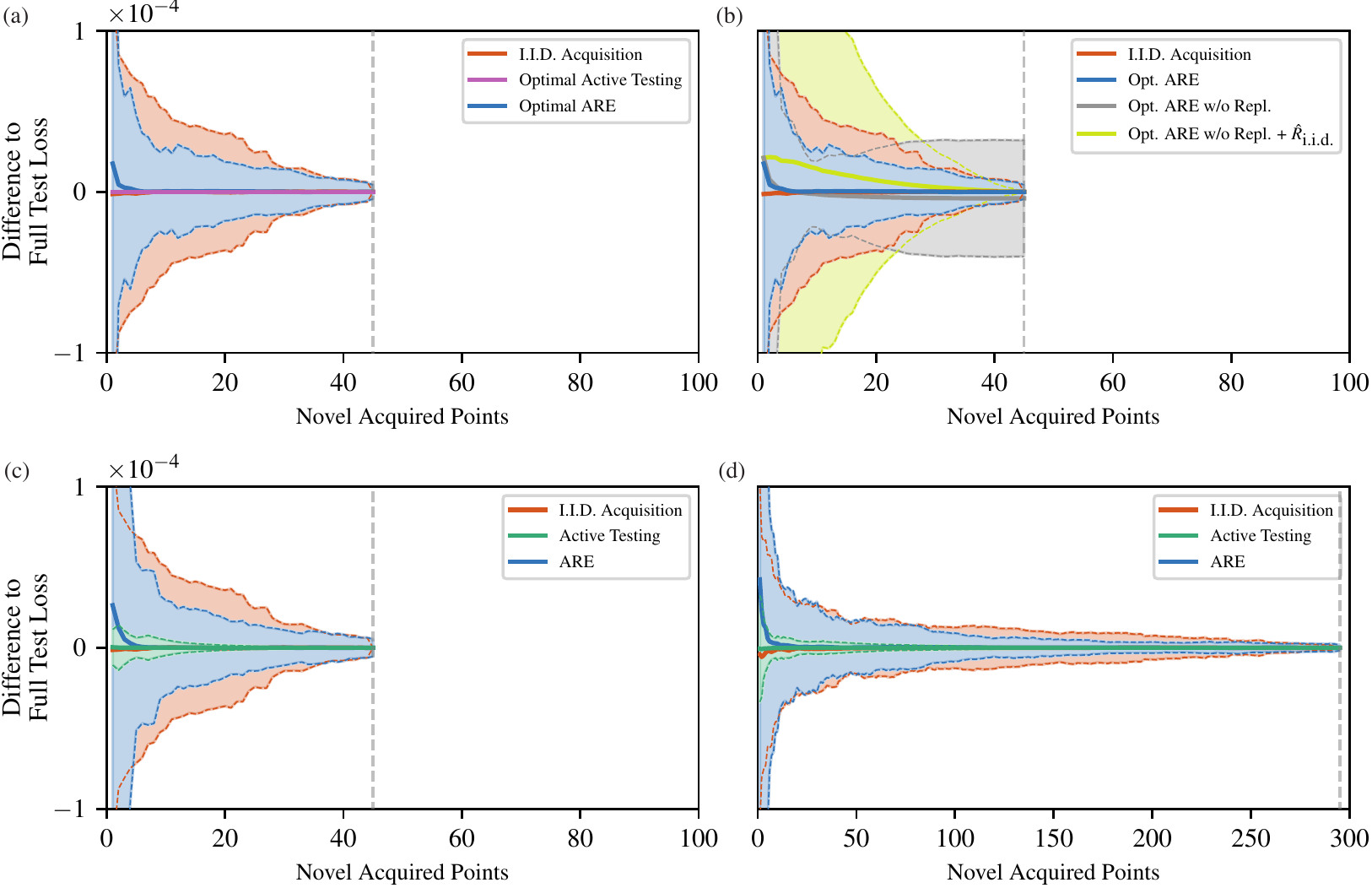}
    \caption{%
    Comparison of Active Testing and Active Risk Estimation.
    Identical to \cref{fig:sawade}, except that now, acquisition steps for ARE are only counted if ARE queries lead to novel acquisitions.
    Displaying results this way slightly improves ARE's performance but does not change our conclusions.
    }
    \label{fig:sawade_novel}
\end{figure}

As \cref{fig:sawade}~(a) shows, `optimal ARE' can improve over i.i.d.~acquisition but does not show the same desirable single-sample optimal behavior as active testing.
Note that ARE extends beyond the test set size of \num{45} because it does not naturally converge due to sampling with replacement.
In \cref{fig:sawade}~(b) we confirm that naive extensions of ARE to sampling \emph{without replacement} are not successful:
(i)~Simply using ARE but sampling without replacement does not work;
(ii)~When additionally using $\Riid$ instead of $\hat{R}_\textup{IS}$, we converge to the empirical test risk but observe a bias and increased variance over i.i.d.~acquisition at earlier acquisition steps.

\textbf{Performance in Practice.}
We now investigate behavior of ARE in practice---where the oracle risk is not available and we rely on \eqref{eq:are_optimal_approx} instead.
\Cref{fig:sawade}~(c) shows that while ARE can increase sample-efficiency over i.i.d.~acquisition in practice, it is clearly outperformed by active testing.
For \cref{fig:sawade}~(d), we increase the size of the test set from \num{45} to \num{295}.
We suspect that further increases in test set size will shrink the gap between ARE and active testing without surrogates, because sampling with replacement should become less important.
However, still, active testing clearly outperforms ARE: active testing also varies substantially in the acquisition strategy it uses (in particular allowing the use of surrogates), with these result suggesting its approach is clearly preferable.%

\textbf{Redefining Acquisitions Steps.}
Sampling with replacement is a sub-optimal strategy for methods applied in pool-based settings.
As mentioned above, sampling with replacement in ARE leads to the same samples being acquired multiple times.
For maximum fairness, we now only increase the `acquisition step' counter by one if the sampling with replacement process leads to an acquisition that has not been previously made, i.e.~if the acquisition is \emph{novel}.
This is somewhat justified in practice because we assume that acquiring novel labels is much more expensive than acquisition function evaluations.

\sisetup{separate-uncertainty=true}
However, find that ARE leads to a significant increase in queries compared to active testing\footnote{%
For \cref{fig:sawade}~(c), ARE takes \SI{9.8 \pm 3.1} as many queries as there are samples in the test pool.
This number will increase for larger test pools and shows that sampling with replacement is not a desirable strategy for pool-based active model evaulation.} and still performs far worse even by this beneficial metric.
Namely, we display the results with `rescaled x-axis' in \cref{fig:sawade_novel}.
While the performance of ARE appears slightly improved, our overall conclusions are unaffected.

\end{document}